\documentclass{article}

     \usepackage[final]{neurips_2020}

\usepackage[utf8]{inputenc} 
\usepackage[T1]{fontenc}    
\usepackage{hyperref}       
\usepackage{url}            
\usepackage{booktabs}       
\usepackage{amsfonts}       
\usepackage{nicefrac}       
\usepackage{microtype}      

\usepackage{epsfig}
\usepackage{graphicx}
\usepackage{amsmath}
\usepackage{amssymb}
\usepackage{booktabs}
\usepackage[table]{xcolor}
\usepackage{caption}
\usepackage{subcaption}
\usepackage{setspace}
\usepackage{bm}
\usepackage{gensymb}
\usepackage{multicol}
\usepackage{wrapfig}
\usepackage{float}

\usepackage{array}

\usepackage{color,soul}

\newcommand{\rulesep}{\unskip\ \vrule width 0.75pt\ }

\definecolor{partcolor}{RGB}{171, 50, 57}
\definecolor{citecol}{RGB}{49, 116, 162}
\definecolor{gray75}{gray}{0.65}
\definecolor{col2}{RGB}{224,56,189}
\definecolor{col3}{RGB}{0, 211, 234}
\definecolor{col1}{RGB}{234,180,160}

\usepackage[T1]{fontenc}
\usepackage[utf8]{inputenc}

\usepackage{natbib}
\setcitestyle{authoryear}

\title{AViD Dataset: Anonymized Videos from Diverse Countries}

\author{%
  AJ Piergiovanni \\
  Indiana University \\
  \texttt{ajpiergi@indiana.edu} \\
  \And
  Michael S. Ryoo \\
  Stony Brook University \\
  \texttt{mryoo@cs.stonybrook.edu} \\
}

\begin{document}

\maketitle

\begin{abstract}
  We introduce a new public video dataset for action recognition: Anonymized Videos from Diverse countries (AViD). Unlike existing public video datasets, AViD is a collection of action videos from many different countries. The motivation is to create a public dataset that would benefit training and pretraining of action recognition models for everybody, rather than making it useful for limited countries. Further, all the face identities in the AViD videos are properly anonymized to protect their privacy. It also is a static dataset where each video is licensed with the creative commons license. We confirm that most of the existing video datasets are statistically biased to only capture action videos from a limited number of countries. We experimentally illustrate that models trained with such biased datasets do not transfer perfectly to action videos from the other countries, and show that AViD addresses such problem. 
  We also confirm that the new AViD dataset could serve as a good dataset for pretraining the models, performing comparably or better than prior datasets\footnote{The dataset is available \href{https://github.com/piergiaj/AViD}{https://github.com/piergiaj/AViD}}.
\end{abstract}

\section{Introduction}
Video recognition is an important problem with many potential applications. One key challenge in training a video model (e.g., 3D spatio-temporal convolutional neural networks) is the lack of data, as these models generally have more parameters than image models requiring even more data. Kinetics \citep{kay2017kinetics} found that by training on a hundreds of thousands of labeled video clips, one is able to increase the performance of video models significantly. Other large-scale datasets, such as HVU \citep{hvu}, Moments-in-Time \citep{monfort2018moments}, and HACS \citep{zhao2019hacs} also have been introduced, motivated by such findings.

\begin{figure}
    \centering
    \begin{subfigure}{0.18\linewidth}
        \centering
        \includegraphics[width=.99\linewidth,height=2cm]{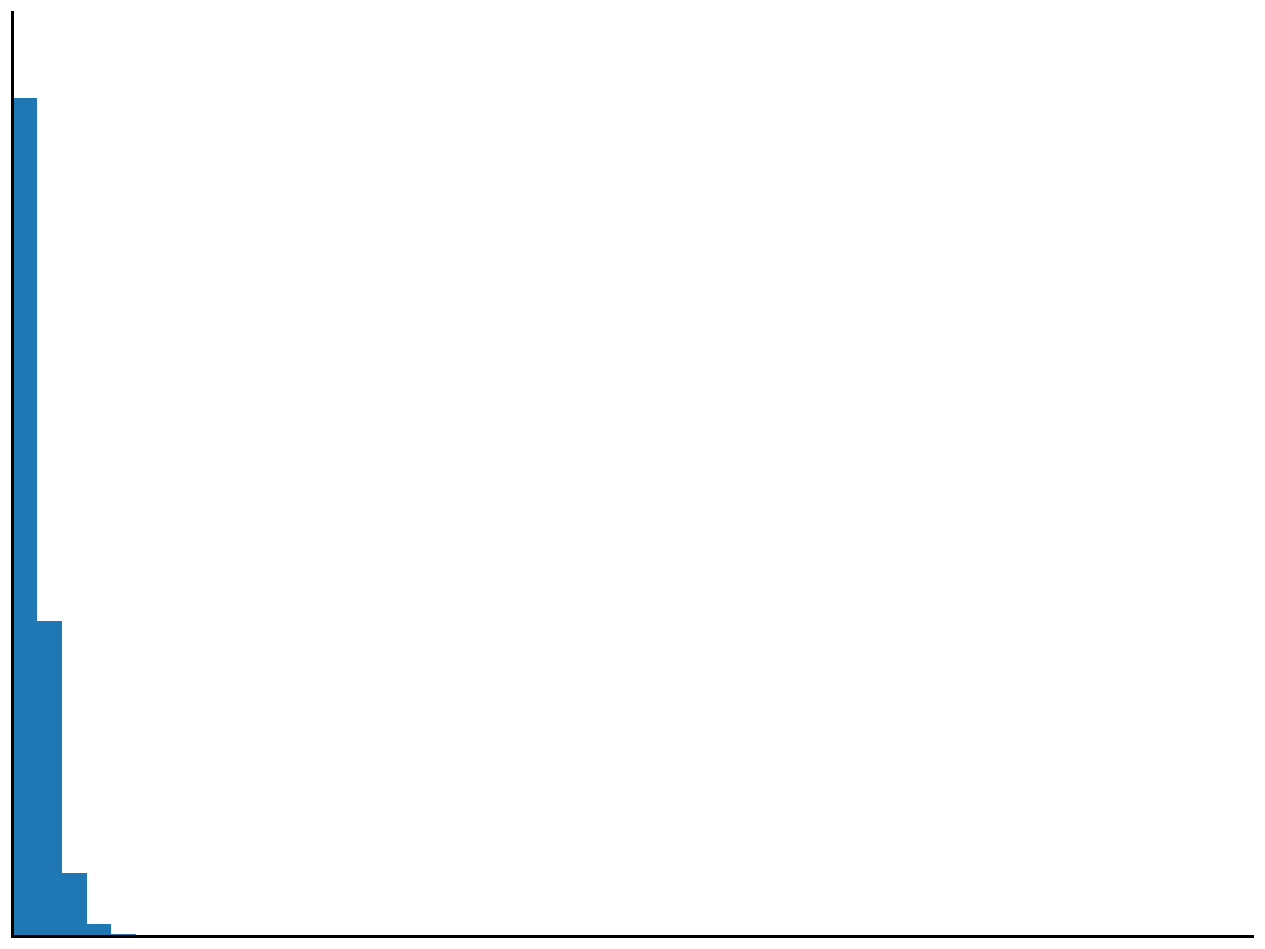}
    \end{subfigure}\rulesep%
    \begin{subfigure}{0.18\linewidth}
        \centering
        \includegraphics[width=.99\linewidth,height=2cm]{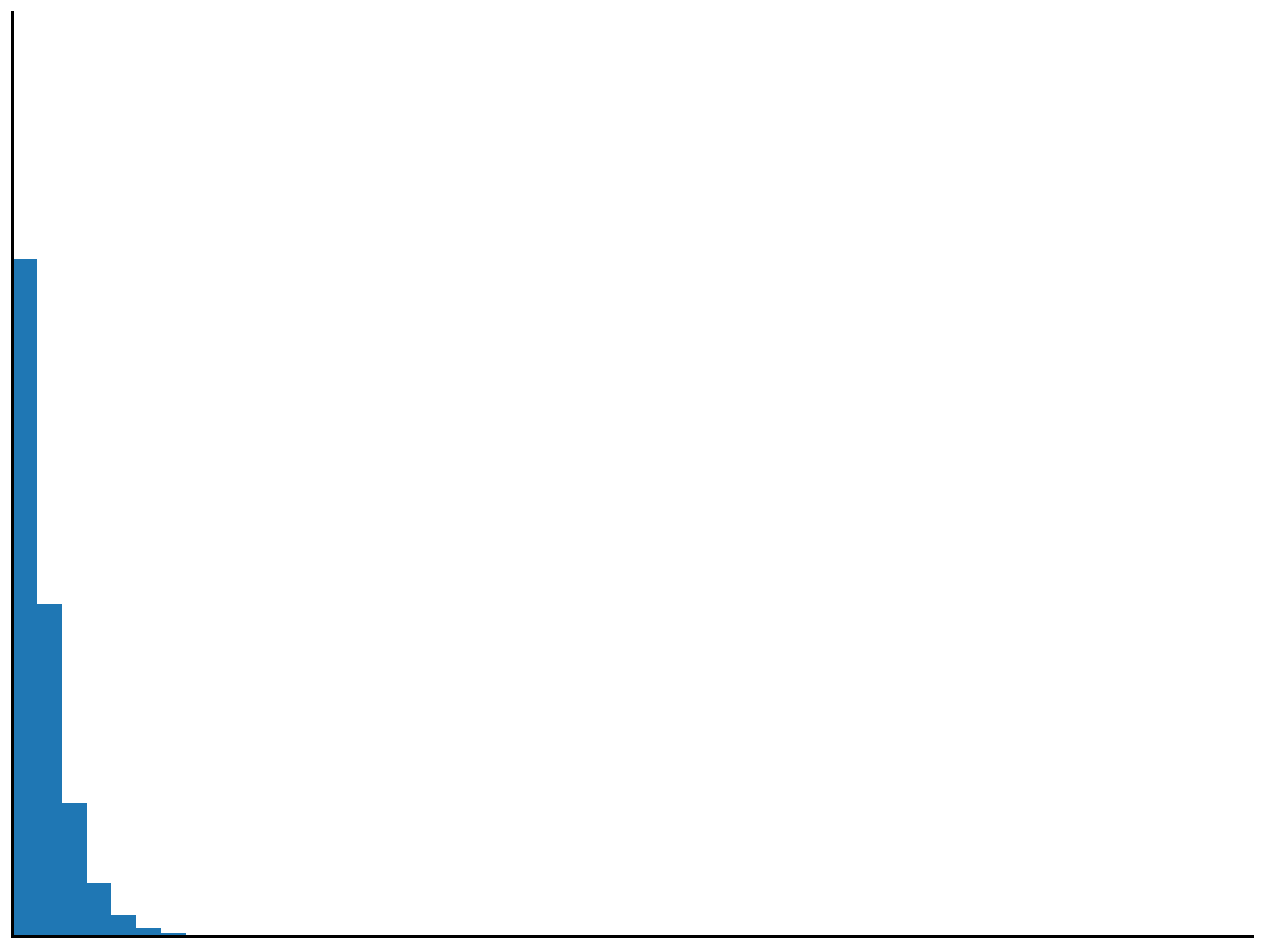}
    \end{subfigure}\rulesep%
    \begin{subfigure}{0.18\linewidth}
        \centering
        \includegraphics[width=.99\linewidth,height=2cm]{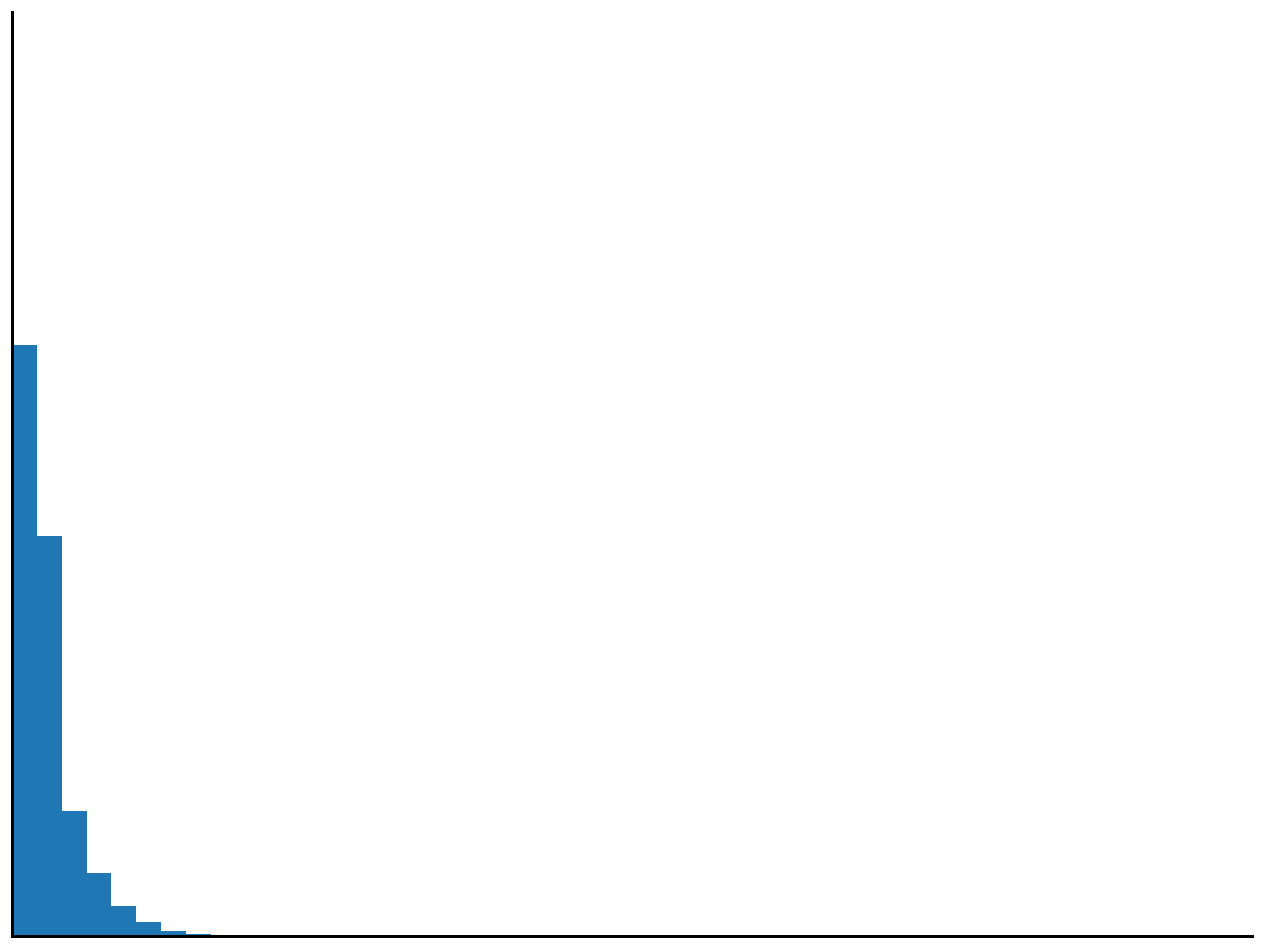}
    \end{subfigure}\rulesep%
    \begin{subfigure}{0.18\linewidth}
        \centering
        \includegraphics[width=.99\linewidth,height=2cm]{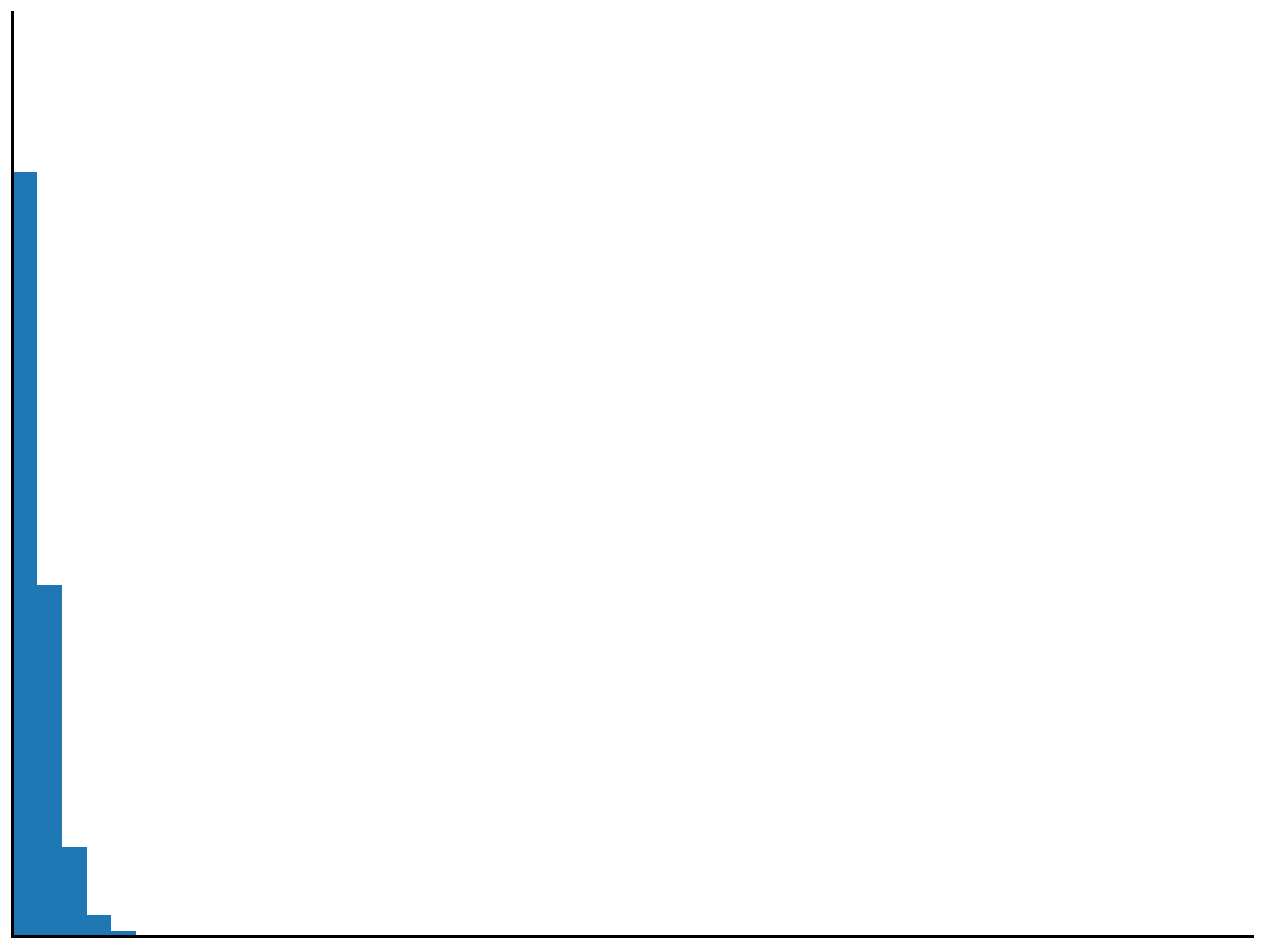}
    \end{subfigure}\rulesep%
    \begin{subfigure}{0.18\linewidth}
        \centering
        \includegraphics[width=.99\linewidth,height=2cm]{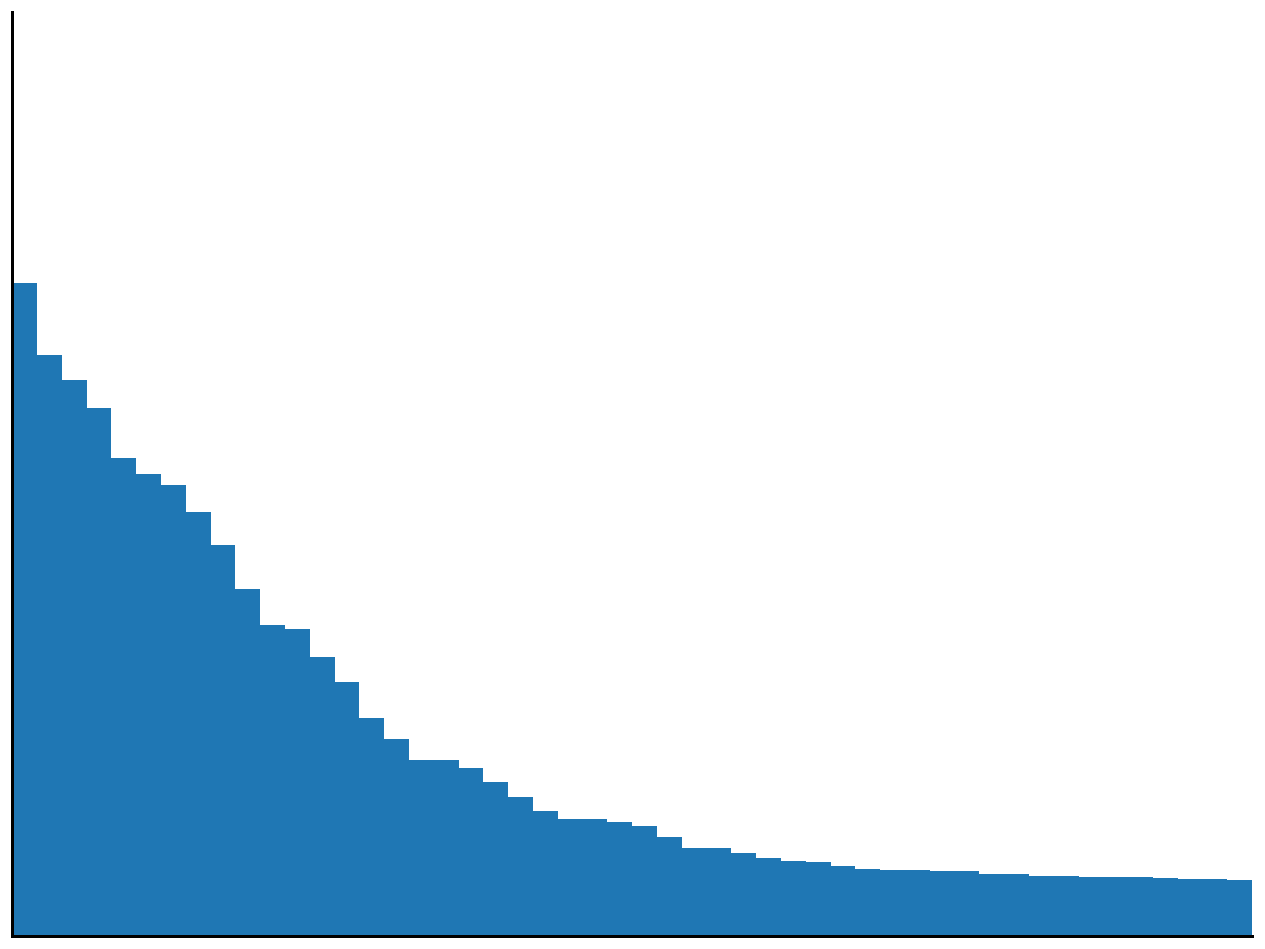}
    \end{subfigure}
    
    \begin{subfigure}{0.18\linewidth}
        \centering
        \includegraphics[width=.99\linewidth]{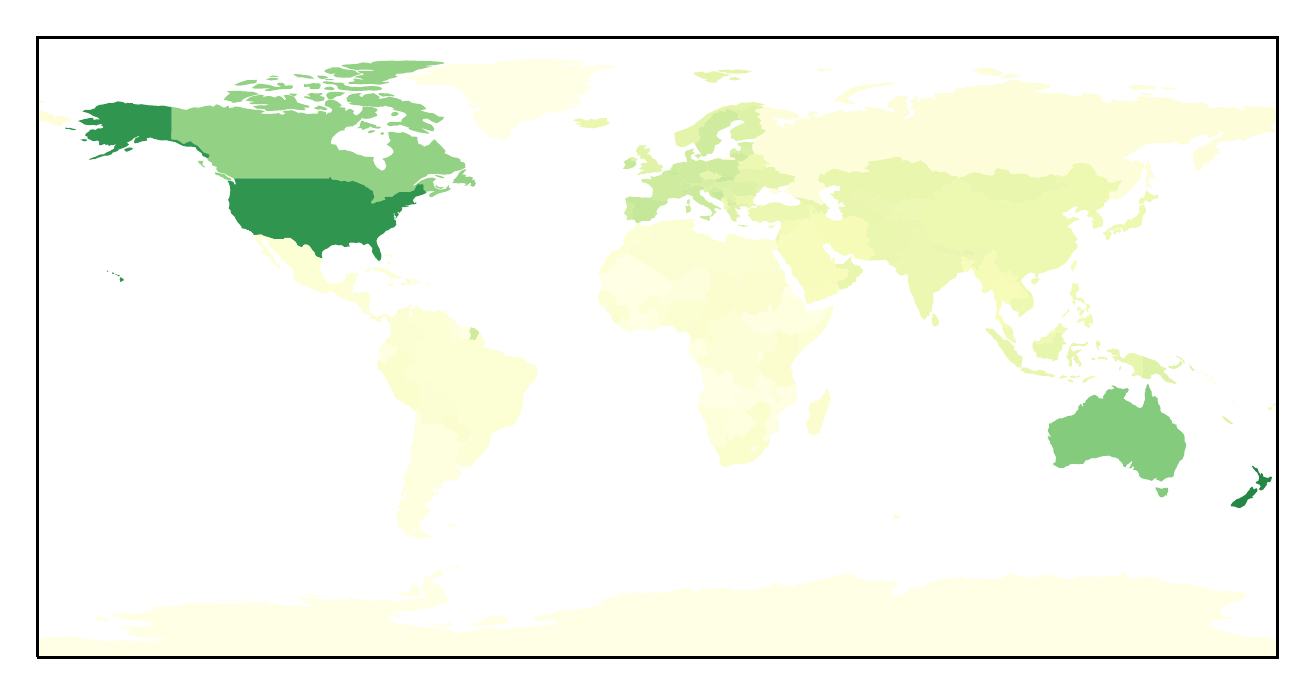}
         \caption{Kinetics-400}
         \label{fig:map-k4}
    \end{subfigure}\rulesep%
    \begin{subfigure}{0.18\linewidth}
        \centering
        \includegraphics[width=.99\linewidth]{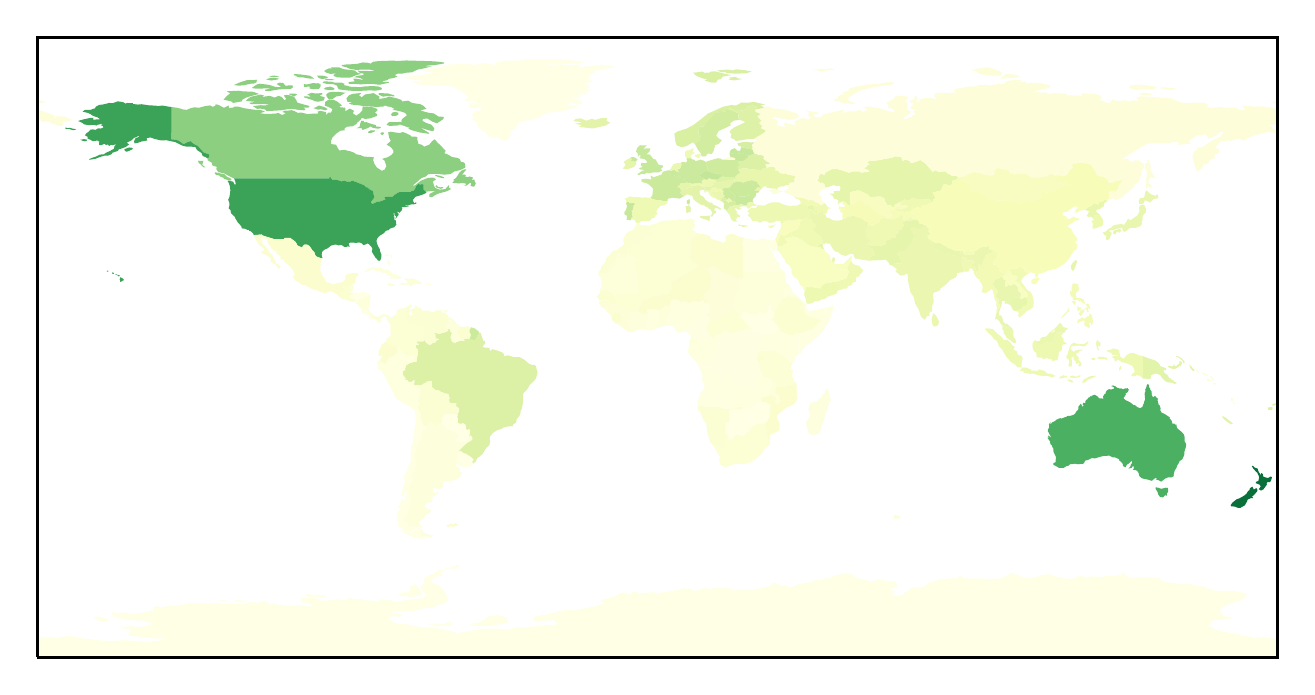}
         \caption{Kinetics-600}
         \label{fig:map-k6}
    \end{subfigure}\rulesep%
    \begin{subfigure}{0.18\linewidth}
        \centering
        \includegraphics[width=.99\linewidth]{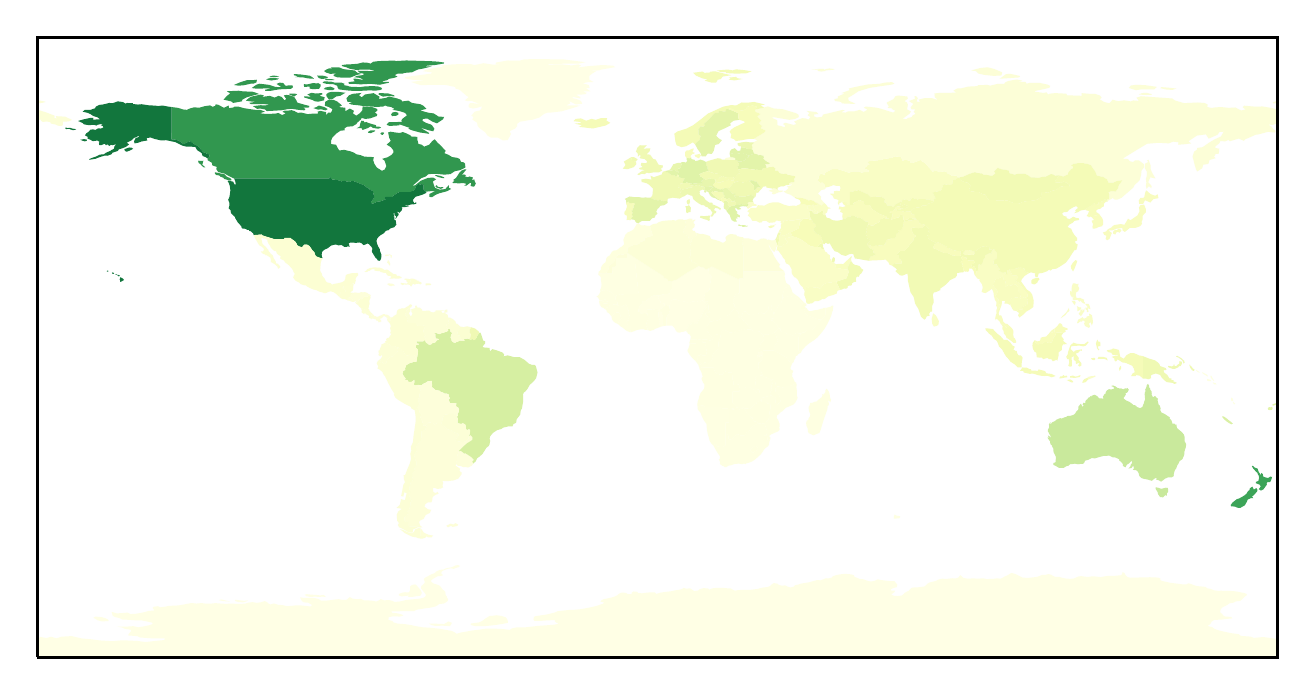}
         \caption{HVU}
         \label{fig:map-hvu}
    \end{subfigure}\rulesep%
    \begin{subfigure}{0.18\linewidth}
        \centering
        \includegraphics[width=.99\linewidth]{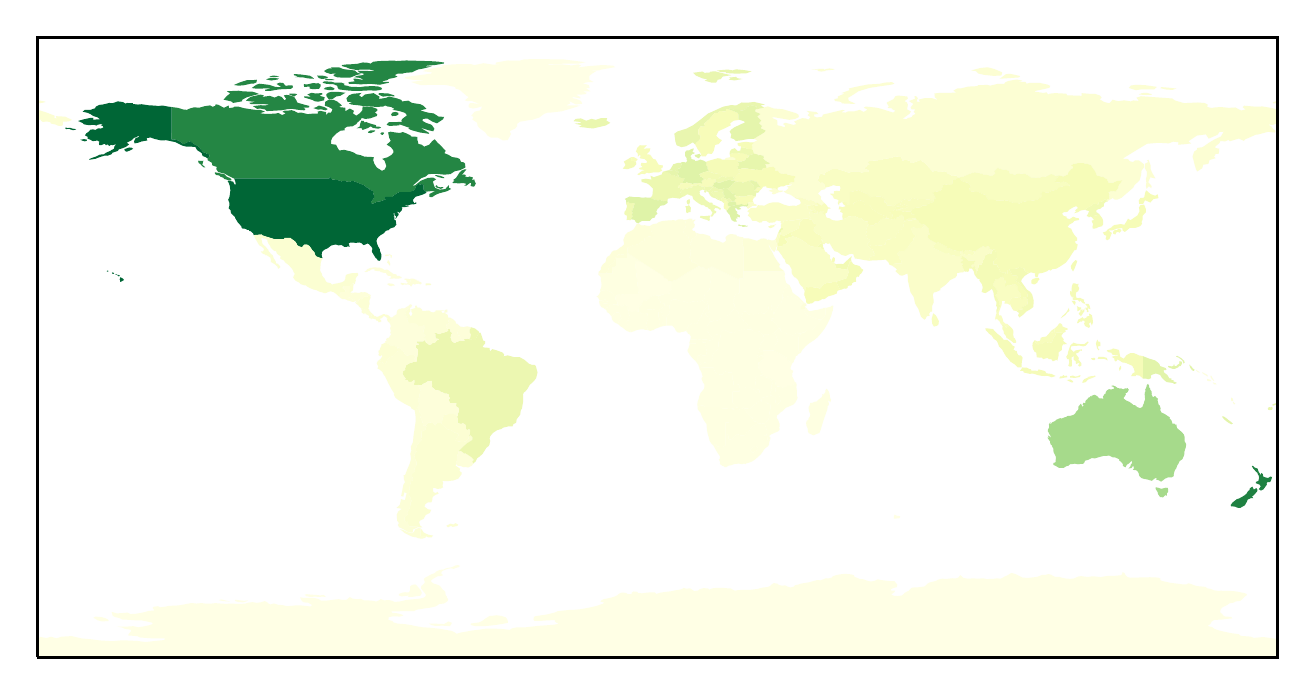}
         \caption{HACS}
         \label{fig:map-hacs}
    \end{subfigure}\rulesep%
    \begin{subfigure}{0.18\linewidth}
        \centering
        \includegraphics[width=.99\linewidth]{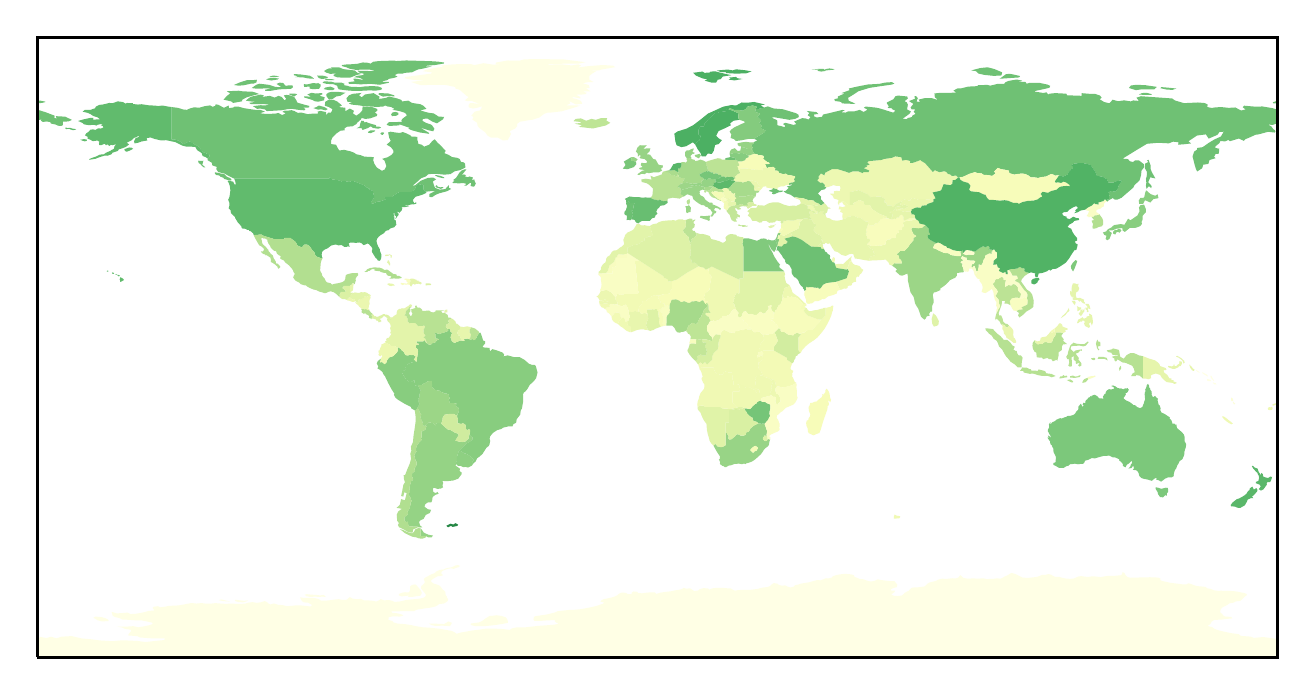}
         \caption{AViD}
         \label{fig:map-avid}
    \end{subfigure}
    \caption{Histogram and Heatmap describing geological distributions of videos for Kinetics and AViD. Video locations are obtained from their geotags using the public YouTube API (check Appendix for details). X-axis of the above histogram correspond to different countries and Y-axis correspond to the number of videos. The color in heatmap is proportional to the number of videos from each country. Darker color means more videos. As shown, AViD has more diverse videos than the others. 
    }
    \label{fig:country-hist}    \vspace{-3mm}
\end{figure}

\begin{wrapfigure}{r}{0.4\linewidth}
    \centering
    \vspace{-0.6cm}
    \includegraphics[width=\linewidth]{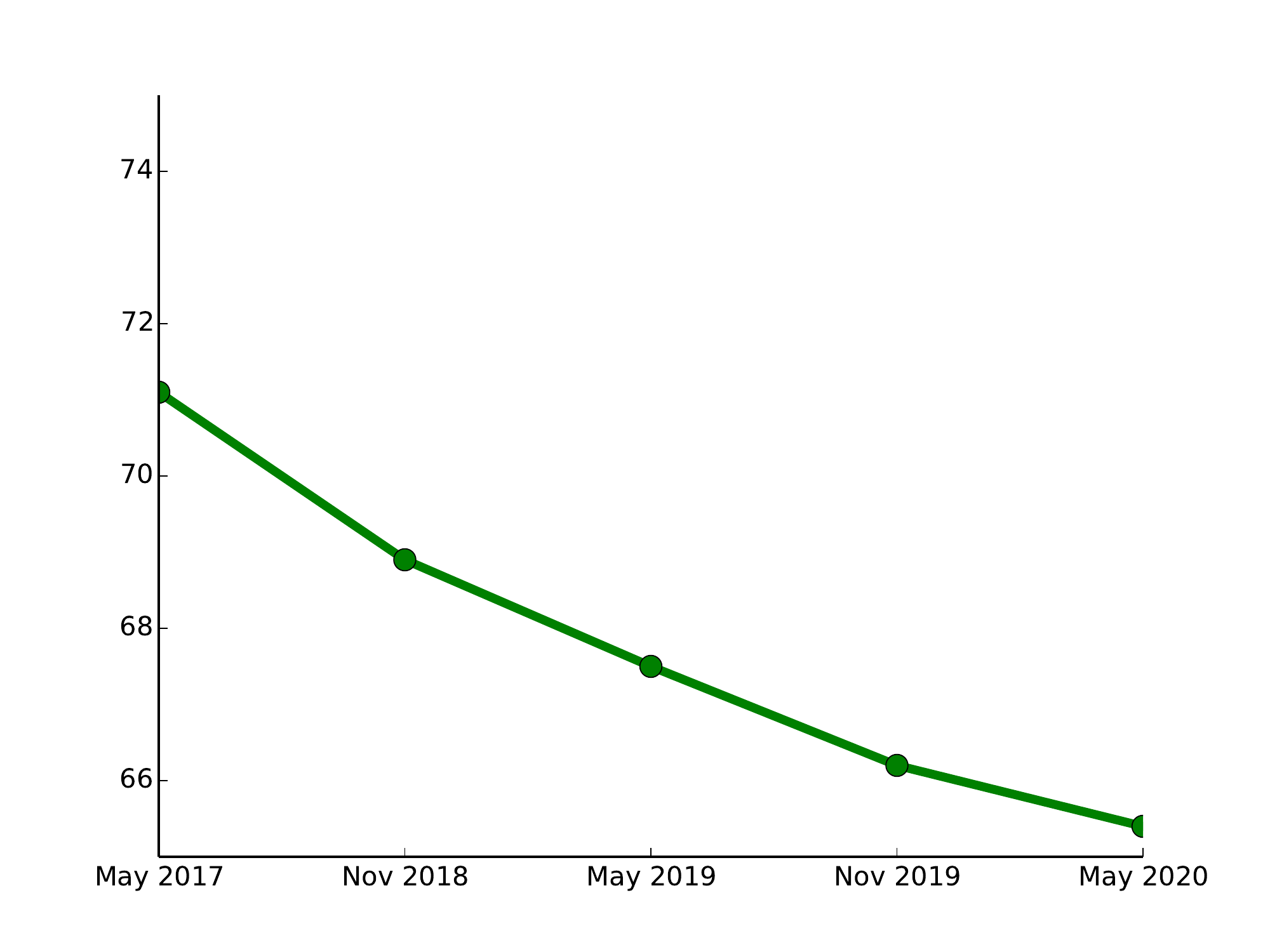}
    \caption{Performance of Kinetics-400 over time as more videos are removed from YouTube. The performance is constantly dropping.}
    \label{fig:kinetics}
    \vspace{-0.3cm}
\end{wrapfigure}

However, many of today's large-scale datasets suffer from multiple problems:
First, due to their collection process, the videos in the datasets are very biased particularly in terms of where the videos are from (Fig. \ref{fig:country-hist} and Table \ref{tab:diverse-stats}).
Secondly, many of these datasets become inconsistent as YouTube videos get deleted. For instance, in the years since Kinetics-400 was first released, over 10\% of the videos have been removed from YouTube. Further, depending on geographic location, some videos may not be available. 
This makes it very challenging for researchers in different countries and at different times to equally benefit from the data and reproduce the results, making the trained models to be biased based on when and where they were trained. They are not static datasets (Figure \ref{fig:kinetics}).


\begin{figure}
    \centering
    \includegraphics[width=\linewidth]{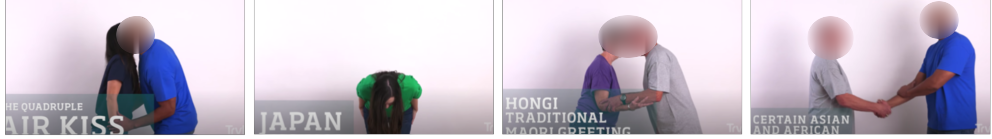}
    \caption{Examples of `greeting' in four different countries. Without diverse videos from all over the world, many of these would not be labeled as `greeting' by a model. These examples are actual video frames from the AViD dataset.}
    \label{fig:greetings}
    \vspace{-2mm}
\end{figure}

AViD, unlike previous datasets, contains videos from diverse groups of people all over the world. Existing datasets, such as Kinetics, have videos mostly from from North America \citep{kay2017kinetics} due to being sampled from YouTube and English queries. AViD videos are distributed more broadly across the globe (Fig. \ref{fig:country-hist}) since they are sampled from many sites using many different languages. This is important as certain actions are done differently in different cultures, such as greetings (shown in Fig. \ref{fig:greetings}), nodding, etc. As many videos contain text, such as news broadcasts, the lack of diversity can further bias results to rely on English text which may not be present in videos from different regions of the world. Experimentally, we show diversity and lack of diversity affects the recognition. 

Further, we anonymize the videos by blurring all the faces. This prevents humans and machines from identifying people in the videos. This is an important property for institutions, research labs, and companies respecting privacy to take advantage the dataset. Due to this fact, face-based actions (e.g., smile, makeup, brush teeth, etc.) have to be removed as they would be very difficult to recognize with blurring, but we show that the other actions are still reliably recognized.

Another technical limitation with YouTube-based datasets including Kinetics, ActivityNet \citep{caba2015activitynet}, YouTube-8M \citep{abu2016youtube}, HowTo100M \citep{miech2019howto100m}, AVA \citep{ava2017} and others, is that downloading videos from YouTube is often blocked. The standard tools for downloading videos can run into request errors (many issues on GitHub exist, with no permanent solution). These factors limit many researchers from being able to use large-scale video datasets.

To address these challenges, we introduce a new, large-scale dataset designed to solve these problems. The key benefits of this dataset is that it captures the same actions as Kinetics plus hundreds of new ones. Further, we choose videos from a variety of sources (Flickr, Instagram, etc.) that have a creative-commons licence. This license allows us to download, modify and distribute the videos as needed. We create a \textbf{static} video dataset that can easily be downloaded. We further provide tags based on the user-generated tags for the video, enabling studying of weakly-labeled data learning. Also unique is the ability to add `no action' which we show helps in action localization tasks. To summarize,
\begin{itemize}
  \setlength\itemsep{1mm}
\item AViD contains actions from diverse countries obtained by querying with many languages.
\item AViD is a dataset with face identities removed
\item AViD is a static dataset with all the videos having the creative-commons licence.
\end{itemize}

\section{Dataset Creation}
The dataset creation process follows multiple steps. First we generated a set of action classes. Next, we sampled videos from a variety of sources to obtain a diverse sample of all actions. Then we generate candidate clips from each video. These clips are then annotated by human. We now provide more details about this process.

\subsection{Action Classes}
Unlike images, where objects are clearly defined and have physical boundaries, determining an action is in videos is a far more ambiguous task. In AViD, we follow many previous works such as Kinetics \citep{kay2017kinetics}, where an action consists of a verb and a noun when needed. For example, `cutting apples' is an action with both a verb and noun while `digging' is just verb.

To create the AViD datasets, the action classes begin by combining the actions in Kinetics, Charades, and Moments in Time, as these cover a wide variety of possible actions. We then remove all actions involving the face (e.g., `smiling,' `eyeliner,' etc.) since we are blurring faces, as this makes it extremely difficult to recognize these actions. Note that we do leave actions like `burping' or `eating' which can be recognized by other contextual cues and motion. 
We then manually combine duplicate/similar actions. This resulted in a set of 736 actions. During the manual annotation process, we allowed users to provide a text description of the actions in the video if none of the candidate actions were suitable and the additional `no action' if there was no action in the video. Based on this process, we found another 159 actions, resulting in 887 total actions. Examples of some of the new ones are `medical procedures,' `gardening,' `gokarting,' etc.

Previous works have studied using different forms of actions, some finding actions associated with nouns to be better \citep{whatactions} while others prefer atomic, generic action \citep{ava2017}. The Moments in Time \citep{monfort2018moments} takes the most common verbs to use as actions, while Charades \citep{sigurdsson2016hollywood} uses a verb and noun to describe each action. Our choice of action closely follows these, and we further build a hierarchy that will enable studying of verb-only actions compared to verb+noun actions and levels of fine-grained recognition.

\subsubsection{Hierarchy}
After deciding the action classes, we realized there was a noticeable hierarchy capturing these different actions. Hierarchies have been created for ImageNet \citep{imagenet_cvpr09} to represent relationships such as fine-grained image classification, but they have not been widely used in video understanding. ActivityNet \citep{caba2015activitynet} has a hierarchy, but is a smaller dataset and the hierarchy mostly capture broad differences and only has 200 action classes.

We introduce a hierarchy that captures more interesting relationships between actions, such as `fishing' $\rightarrow$ `fly tying,' `casting fishing line,' `catching fish,' etc. And more broad differences such as `ice fishing' and `recreational fishing.' Similarly, in the `cooking class' we have `cutting fruit' which has both `cutting apples' and `cutting pineapple'. Some actions, like `cutting strawberries' didn't provide enough clips (e.g., less than 10), and in such case, we did not create the action category and made the videos only belong to the `cutting fruit' class. This hierarchy provides a starting point to study various aspects of what an action is, and how we should define actions and use the hierarchy in classifiers. Part of the  hierarchy is shown in Fig. \ref{fig:hier}, the full hierarchy is provided in the supplementary material.

\begin{figure}
    \centering
    \includegraphics[width=\linewidth]{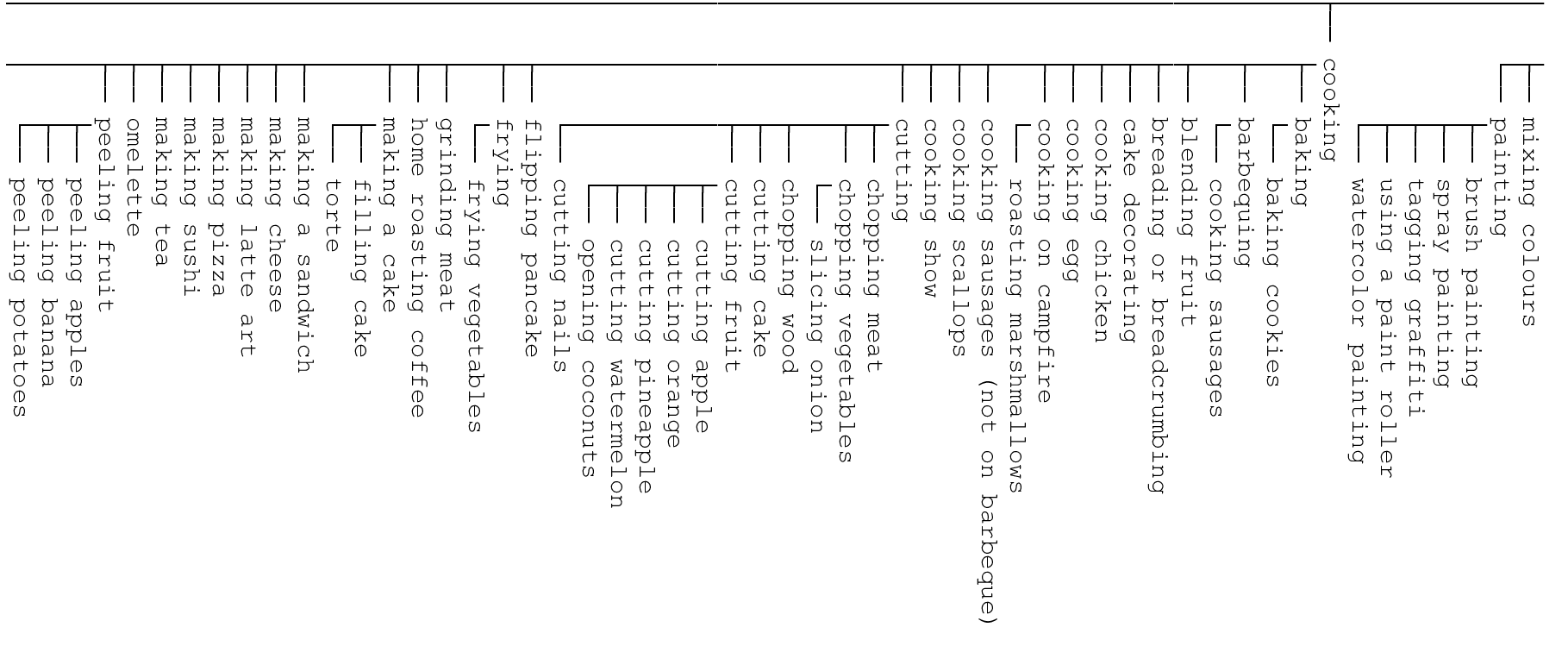}
    \caption{Illustration of a section of the hierarchy of activities in AViD. Check Appendix for the full hierarchy with 887 classes.}
    \label{fig:hier}
\end{figure}

\subsection{Video Collection}
AViD videos are collected from several websites: Flickr, Instagram, etc. But we ensure all videos are licensed with the creative commons license. This allows us to download, modify (blur faces), and distribute the videos. This enables the construction of a static, anonymized, easily downloadable video dataset for reproducible research.

In order to collect a \emph{diverse} set of candidate videos to have in the dataset, we translated the initial action categories into 22 different languages (e.g., English, Spanish, Portuguese, Chinese, Japanese, Afrikaans, Swahili, Hindi, etc.) covering every continent. We then searched multiple video websites (Instagram, Flickr, Youku, etc.) for these actions to obtain initial video samples. This process resulted in a set of 800k videos. From these videos, we took multiple sample clips. As shown in Fig. \ref{fig:country-hist}, this process found videos from all over the globe. 

We ensured there was no overlap of AViD videos and those in the validation or testing sets of Kinetics. There is some minor overlap between some of AViD videos and the training set of Kinetics, which is an outcome due to that the both datasets were collected from the web. 

\subsection{Action Annotation}
We annotate the candidate clips using Amazon Mechanical Turk. In order to make human annotations more efficient, we use I3D model \citep{carreira2017quo} to generate a set of potential candidate labels for each clip (the exact number depends on how many actions I3D predicted, usually 2-3) and provide them as suggestions to the human annotators. We also provide annotators an option to select the `other' and `none' category and manually specify what the action is. For each task, one of the videos was from an existing dataset where the label was known. This served as a quality check and the annotations were rejected if the worker did not correctly annotate the test video. A subset of the videos where I3D (trained with Kinetics) had very high confidence (> 90\%) were verified manually by the authors. 

As a result, a total of 500k video clips were annotated. Human annotators labeled 300k videos manually, and 200k videos with very high-confidence I3D predictions were checked by the authors and the turkers.
Of these, about 100k videos were labeled as the `other' action by the human annotators, suggesting that I3D + Kinetics training does not perform well on these actions. Of these, about 50k videos were discarded due to poor labeling or other errors, resulting in a dataset of 450k total samples.

\begin{figure}
    \centering
    \includegraphics[width=0.5\linewidth]{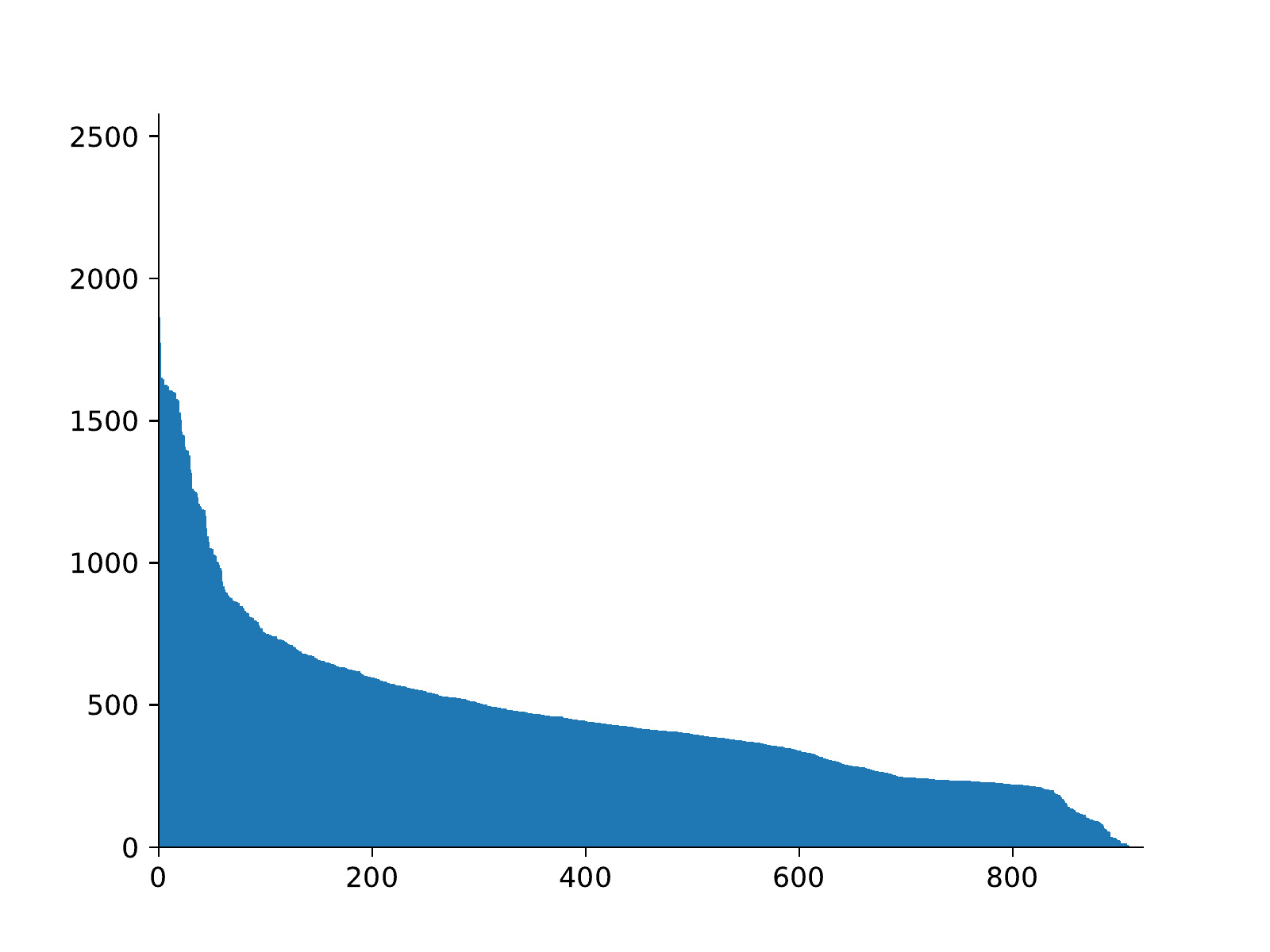}
    \caption{Distribution of videos per class in the AViD dataset. We find it follows a Zipf distribution, similar to the actions in other large-scale video datasets.}
    \label{fig:classdist}
\end{figure}

We found the distribution of actions follows a Zipf distribution (shown in Fig. \ref{fig:classdist}, similar to the observation of AVA \citep{ava2017}. We split the dataset into train/test sets by taking 10\% of each class as the test videos. This preserves the Zipf distribution. 

\begin{table}[]
    \centering
    \caption{Comparison of large video datasets for action classification.}
    \label{tab:dataset-sizes}
    \begin{tabular}{c|ccccc}
    \toprule
        Dataset & Classes & Train Clips & Test Clips &  Hours & Clip Dur.\\
    \midrule
        Kinetics-400 & 400 & 230k & 20k & 695 & 10s \\
        Kinetics-600 & 600 & 392k & 30k &  1172 & 10s \\
        Moments in Time & 339 & 802k & 33k & 667 & 3s\\
        AViD & 887 & 410k & 40k & 880 & 3-15s\\
    \bottomrule
    \end{tabular}
\end{table}

\subsection{Weak Tag Annotation}
In addition to action category annotation per video clips, AviD dataset also provides a set of weak text tags. To generate the weak tags for the videos, we start by translating each tag (provided from the web) into English. We then remove stopwords (e.g., `to,' `the,' `and,' etc.) and lemmatize the words (e.g., `stopping' to `stop'). This transforms each tag into its base English word. 

\begin{figure}
    \centering
    \begin{subfigure}{0.24\linewidth}
        \centering
        \includegraphics[width=.99\linewidth]{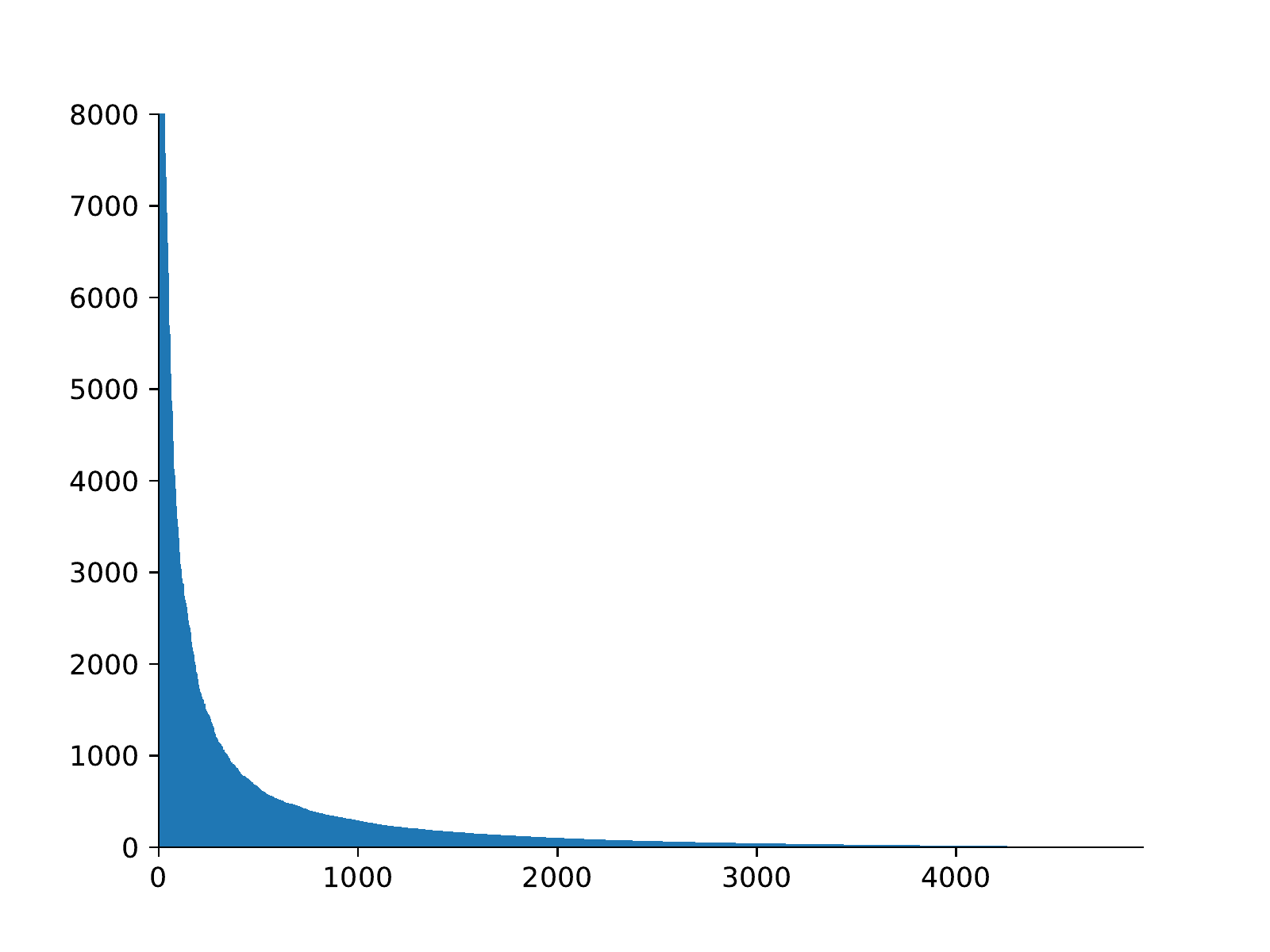}
         \caption{}
         \label{fig:t1d}
    \end{subfigure}%
    \begin{subfigure}{0.24\linewidth}
        \centering
        \includegraphics[width=.99\linewidth]{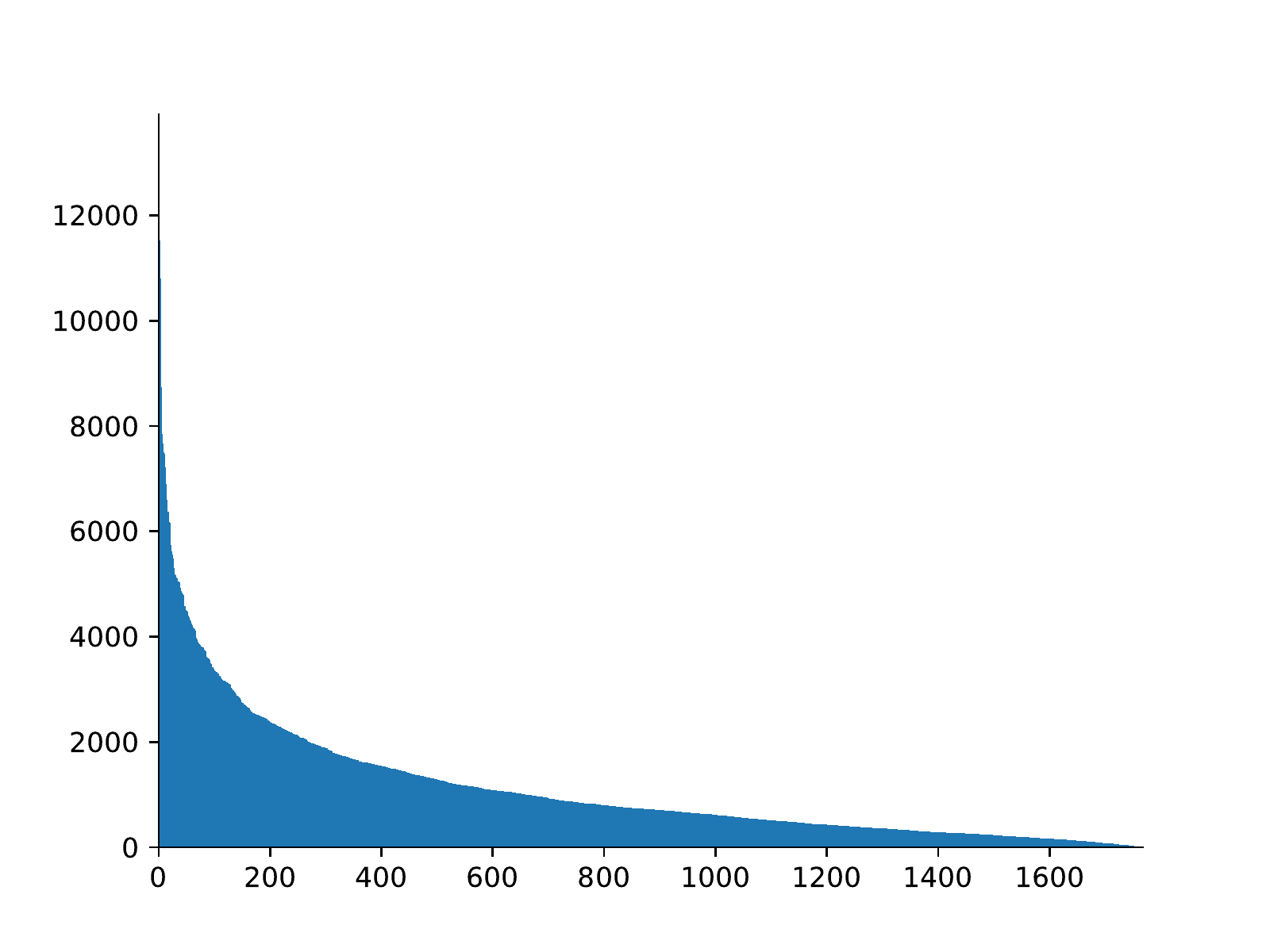}
         \caption{}
         \label{fig:t2d}
    \end{subfigure}\rulesep%
    \begin{subfigure}{0.24\linewidth}
        \centering
        \includegraphics[width=.99\linewidth]{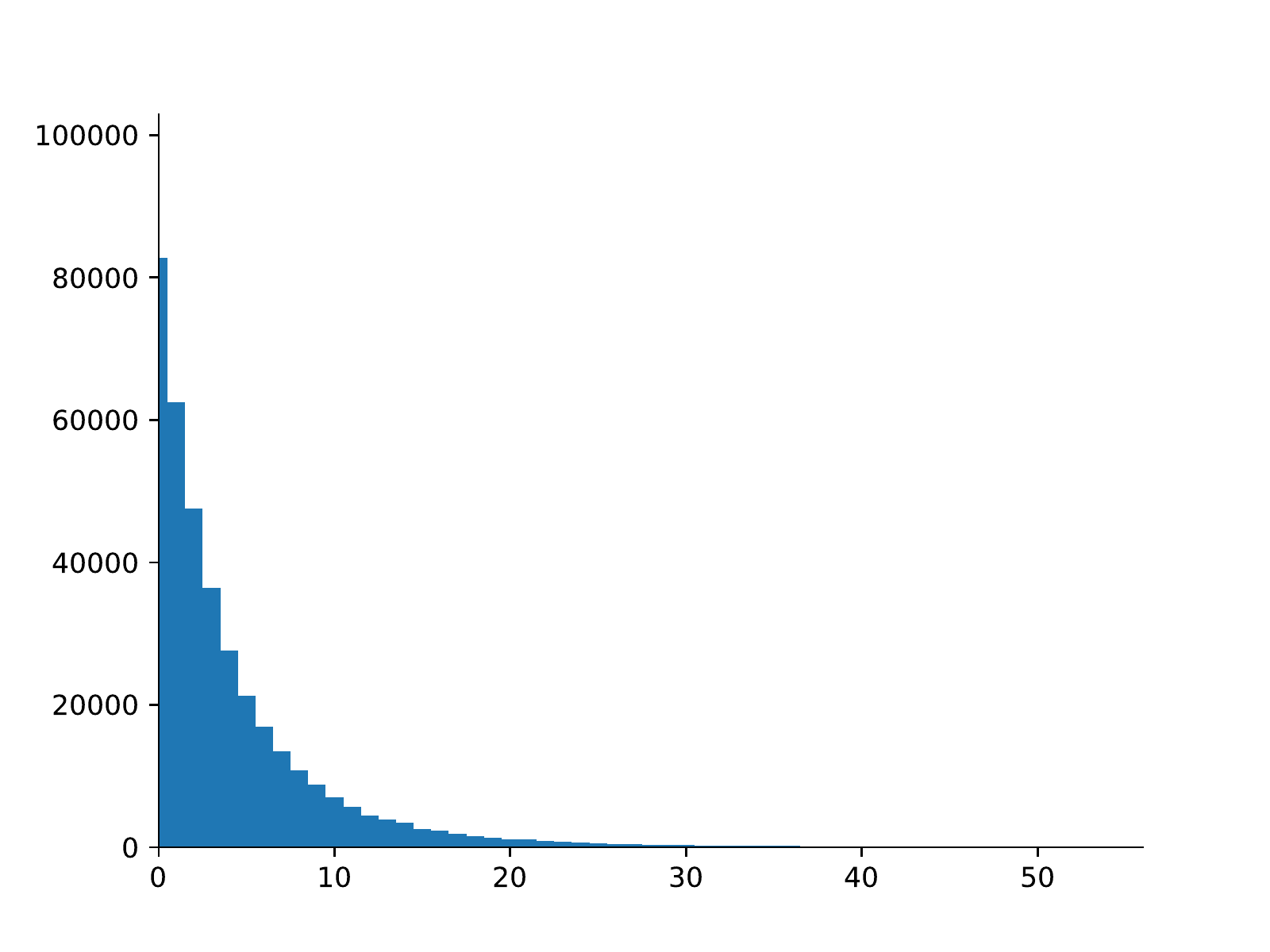}
         \caption{}
         \label{fig:v1}
    \end{subfigure}%
    \begin{subfigure}{0.24\linewidth}
        \centering
        \includegraphics[width=.99\linewidth]{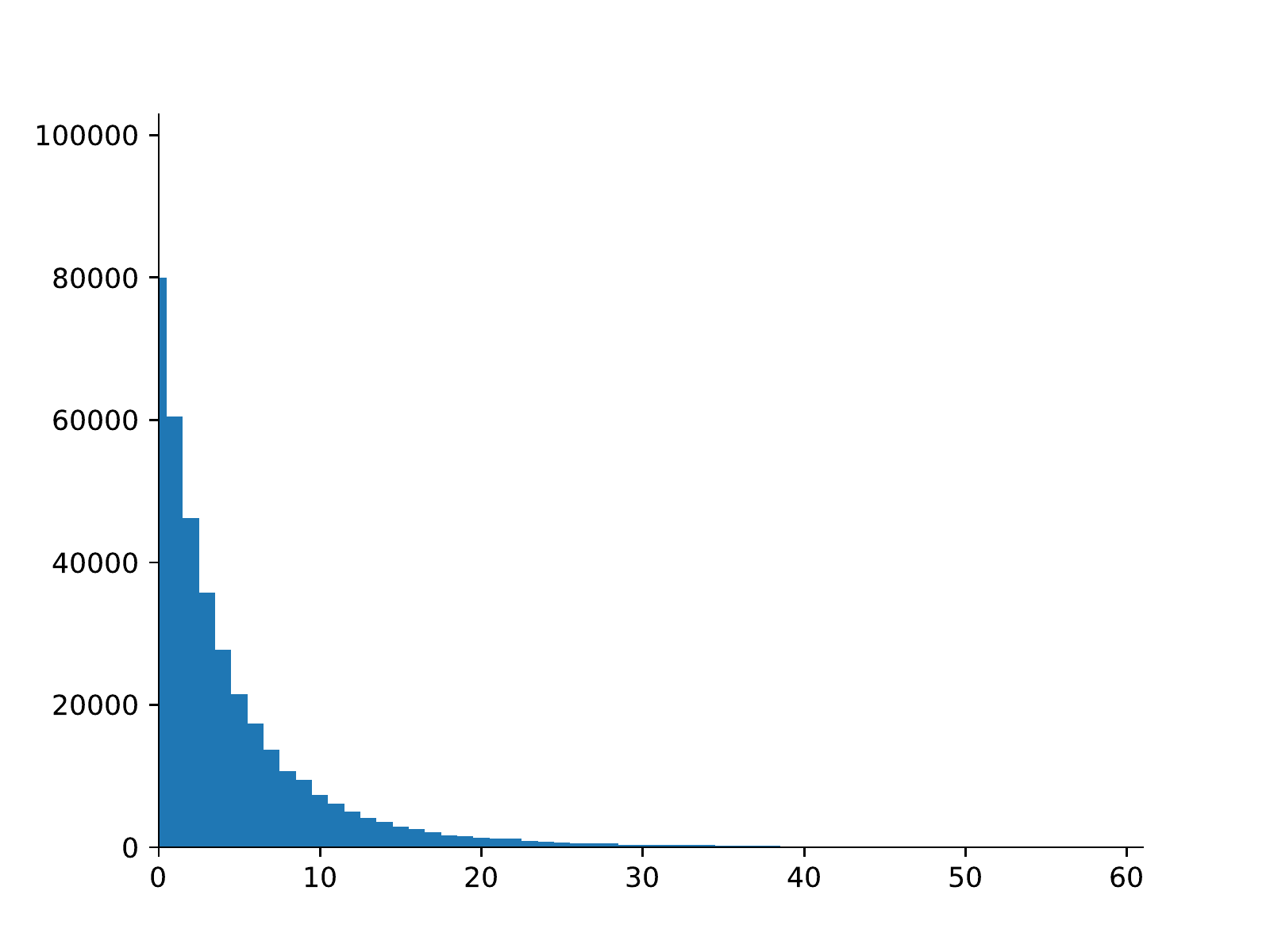}
         \caption{}
         \label{fig:v2}
    \end{subfigure}
    \caption{Evaluation of the weak tag distributions. \textbf{(a/b)} Number of times each tag appears in the dataset from the agglomerative clustering or affinity propagation. \textbf{(c/d)} Number of tags in each video. Videos have between 0 and 65 tags, most have 1-8 tags.}
    \label{fig:tag_dist}
\end{figure}

Next, we use word2vec \citep{word2vec} to compute the distance between each pair of tags, and use affinity propagation and agglomerative clustering to generate 1768 and 4939 clusters, respectively. Each video is then tagged based on these clusters. This results in two different sets of tags for the videos, both of which are provided for further analysis, since it is unclear which tagging strategy will more benefit future approaches. 
The overall distribution of tags is shown in Fig. \ref{fig:tag_dist}, also following an exponential distribution.

\section{Experiments}

We conducted a series of experiments with the new AViD dataset. This not only includes testing existing video CNN models on the AViD dataset and further evaluating effectiveness of the dataset for pretraining, but also includes quantitative analysis comparing different datasets. Specifically, we measure video source statistics to check dataset biases, and experimentally confirm how well a model trained with action videos from biased countries generalize to  videos from different countries. We also evaluate how face blurring influences the classification accuracy, and introduce weak annotations of the dataset.

\paragraph{Implementation Details} We implemented the models in PyTorch and trained them using four Titan V GPUs. To enable faster learning, we followed the multi-grid training schedule \citep{wu2019multigrid}. The models, I3D \citep{carreira2017quo}, 2D/(2+1D)/3D ResNets \citep{he2016deep,tran2018closer,tran2014c3d}, Two-stream \citep{simonyan2014two}, and SlowFast \citep{feichtenhofer2018slowfast}, were trained for 256 epochs. The learning rate followed a cosine decay schedule with a max of 0.1 and a linear warm-up for the first 2k steps. Each GPU used a base batch size of 8 clips, which was then scaled according to the multi-grid schedule (code provided in supplementary materials). The base clip size was 32 frames at $224\times 224$ image resolution.

For evaluation, we compared both convolutional evaluation where the entire $T$ frames at $256\times 256$ were given as input as well as a multi-crop evaluation where 30 random crops of 32 frames at $224\times 224$ are used and the prediction is the average over all clips.

\paragraph{Baseline Results} In Table \ref{tab:baselines}, we report the results of multiple common video model baseline networks. Overall, our findings are consistent with the literature.
\vspace{-2mm}

\begin{table}
    \centering
    \caption{Performance of multiple baselines models on the AViD dataset.}
    \label{tab:baselines}
    \begin{tabular}{c|cc}
         \toprule
         Model & Acc (conv) & Acc (multi-crop)  \\
         \midrule
         2D ResNet-50 & 36.2\% & 35.3\% \\
         I3D \citep{carreira2017quo} & 46.5\% & 46.8\% \\
         3D ResNet-50 & 47.9\% & 48.2\% \\
         Two-Stream 3D ResNet-50 & 49.9\% & 50.1\% \\
         Rep-Flow ResNet-50 \citep{piergiovanni2018representation} & 50.1\% & 50.5\% \\
         (2+1)D ResNet-50 & 46.7\% & 48.8\% \\
         SlowFast-50 4x4 \citep{feichtenhofer2018slowfast} & 48.5\% & 47.4\% \\
         SlowFast-50 8x8 \citep{feichtenhofer2018slowfast} & 50.2\% & 50.4\%\\         
         SlowFast-101 16x8 \citep{feichtenhofer2018slowfast} & 50.8\% & 50.9\% \\
         \bottomrule
    \end{tabular}
\end{table}

\paragraph{Diversity Analysis}

Since AViD is designed to capture various actions from diverse countries, we conduct a set of experiments to measure the diversity and determine the effect of having diverse videos. 

First, we computed geo-location statistics of AViD and other datasets, and compared them. To obtain the locations of AViD videos, we extract the geo-tagged location for videos where it was available (about 75\% of total AViD videos). We used the public API of the site where each AViD video came from to gather the geolocation statistics. Similarly, we used the public YouTube API to gather the geolocation statistics for the Kinetics, HACS, and HVU videos. Further, after the initial release of AViD (on arXiv), the Kinetics team provided us their location statistics estimate \citep{kineticsdiversity}. As it is a bit different from our estimate, we also directly include such data for the comparison.\footnote{We believe the main difference comes from the use of public YouTube API vs. YouTube's internal geolocation metadata estimated based on various factors. Please see the appendix for more details.} 





To measure the diversity of each dataset, we report a few metrics: (1) percentage of videos in North America, Latin America, Europe, Asia, and Africa. (2) As a proxy for diversity and bias, we assume a uniform distribution over all countries would be the most fair (this assumption is debatable), then using the Wasserstein distance, we report the distance from the distribution of videos to the uniform distribution. The results are shown in Table \ref{tab:diverse-stats}. We note that due to the large overlap in videos between HVU and Kinetics-600, their diversity stats are nearly identical. Similarly, as HACS is based on English queries of YouTube, it also results in a highly North American biases dataset. We note that Kinetics-600 and -700 made efforts to improve diversity by querying in Spanish and Portuguese, which did improve diversity in those countries \citep{carreira2018shortkin600,kineticsdiversity}.

\begin{table}[]
    \centering    \caption{Comparing diversity of videos based on geotagged data. The table shows percentages of the videos from North America, Latin American, Europe, Asia, and Africa. `Div' measures the Wasserstein distance between the actual data distribution and the uniform distribution, the lower the more balanced videos are (i.e., no location bias). For Kinetics, we include both our estimated numbers ($^\dagger$) as well as the internal numbers from the Kinetics team \citep{kineticsdiversity}$^2$.}
    \label{tab:diverse-stats}

    \begin{tabular}{c|ccccc|c}
        \toprule
        Dataset & N.A. & L.A. & EU & Asia & AF  & Div  \\
        \midrule
        Kinetics-400$^\dagger$ & 96.2 & 0.3 & 2.3 & 1.1 & 0.1 & 0.284 \\
        Kinetics-400$^2$ & 59.0  & 3.4  & 21.4  & 11.8 & 0.8 & 0.169 \\
        Kinetics-600$^\dagger$ & 87.3 & 6.1 & 4.3 & 2.2 &  0.1&  0.269 \\
        Kinetics-600$^2$ & 59.1  &  5.7 & 19.3 & 11.3 & 0.9 & 0.164 \\
        Kinetics-700$^2$ & 56.8  &  7.6 & 19.6 & 11.5 & 1.0 & 0.158 \\
        HVU & 86.4 & 6.3 & 4.7 & 2.5 &  0.1 &  0.266 \\
        HACS & 91.4 & 1.5 & 5.8 & 1.2 &  0.1 &  0.286 \\
        AViD & 32.5 & 18.6 & 19.7 & 20.5 & 8.7 & 0.052 \\
        \bottomrule
    \end{tabular}
\end{table}

In addition, we ran an experiment training the baseline model on each dataset, and testing it on videos from different regions of the world.
Specifically, we train the baseline 3D ResNet model with either Kinetics-400/600 or AViD. Then we evaluated the models on AViD videos using action classes shared by both Kinetics-400 and AViD (about 397 classes) while splitting evaluation into North American, Rest of World, or other regions. 
The results are summarized in Table \ref{tab:diverse}. We find that the models trained with any of the three datasets perform quite similarly on the North American videos. However, the Kinetics trained models do not perform as well on the diverse videos, while AViD models show a much smaller drop. This suggests that current datasets do not generalize well to diverse world data, showing the importance of building diverse datasets. In Table \ref{tab:diverse2}, we show the results when using all AViD classes, but using training on a specific region then testing on that region vs. all other regions\footnote{There are only $\sim$35k training clips from Africa, and the smaller training set reduces overall performance.}. We observe that the performance drops when training vs. testing are from different regions. This further suggests that having a training set of videos from diverse countries are essential.


\begin{table}
    \centering
    \caption{Effect of having diverse videos during training. Note that we only test on AViD videos with activities shared between Kinetics-400 and AViD (397 classes). We report the accuracy on North American (N.A.) videos and the rest of the world (RoW) videos, and specific region videos.}
    \label{tab:diverse}
    \begin{tabular}{c|cccccccc}
         \toprule
         Model & Training Data & Acc (N.A.) & Acc (RoW) & L.A. & EU & Asia & AF  \\
         \midrule
         3D ResNet-50 & Kin-400 & 72.8\% & 64.5\% & 68.3\% & 71.2\% & 61.5\% & 58.4\% \\
         3D ResNet-50 & Kin-600 & 73.5\% & 65.5\% & 69.3\% & 72.4\% & 62.4\% & 59.4\%\\
         3D ResNet-50 & AViD (all) & 75.2\% & 73.5\% & 74.5\% & 74.3\% & 74.9\%  & 71.4\% \\
    \bottomrule
    \end{tabular}
\end{table}

\begin{table}
    \centering
    \caption{Training on one region and testing on the same and on the others all AViD classes. In all cases, the models perform worse on other regions than the one trained on$^3$. This table uses a 3D ResNet-50.}
    \label{tab:diverse2}
    \begin{tabular}{c|ccc}
         \toprule
          AViD Training Data & Acc (Same Region) & Acc (All Other Regions)  \\
         \midrule
         N.A. & 51.8\% & 42.5\% \\
         L.A. & 49.4\% & 38.5\% \\
         EU & 47.5\% & 39.4\% \\
         Asia & 46.7\% & 41.2\% \\
         Africa$^3$ & 42.5\% & 32.2\% \\
    \bottomrule
    \end{tabular}
\end{table}

\paragraph{Fine-tuning} We pretrain several of the models with AViD dataset, and fine-tune on HMDB-51 \citep{kuehne2011hmdb} and Charades \citep{sigurdsson2016hollywood}.

The objective is to compare AViD with exising datasets in terms of pretraining, including Kinetics-400/600 \citep{kay2017kinetics} and Moments-in-time (MiT) \citep{monfort2018moments}. Note that these results are based on using RGB-only as input; no optical flow is used.

In Table \ref{tab:finetune-hmdb}, we compare the results on HMDB. We find that AViD performs quite similarly to both Kinetics and MiT. Note that the original Kinetics has far more videos than are currently available (as shown in Figure \ref{fig:kinetics}), thus the original fine-tuning performance is higher (indicated in parenthesis).

\begin{table}
    \centering
    \caption{Performance standard models fine-tuned on HMDB. Numbers in parenthesis are based on original, full Kinetics dataset which is no longer available.}
    \label{tab:finetune-hmdb}
    \begin{tabular}{c|ccc}
         \toprule
         Model & Pretrain Data & Acc   \\
         \midrule
         I3D \citep{carreira2017quo} & Kin-400 & 72.5 (74.3) \\
         I3D \citep{carreira2017quo} & Kin-600 & 73.8 (75.4)  \\
         I3D \citep{carreira2017quo} & MiT & 74.7 \\
         I3D \citep{carreira2017quo} & AViD & 75.2 \\
        \midrule
         3D ResNet-50 & Kin-400 & 75.7 (76.7) \\
         3D ResNet-50 & Kin-600 & 76.2 (77.2) \\
         3D ResNet-50 & MiT & 75.4 \\  
         3D ResNet-50 & AViD & 77.3 \\
    \bottomrule
    \end{tabular}
\end{table}

In Table \ref{tab:finetune-charades}, we compare the results on the Charades dataset. Because the AViD dataset also provides videos with `no action' in contrast to MiT and Kinetics which only have action videos, we compare the effect of using `no action' as well. While AViD nearly matches or improves performance even without `no action' videos in the classification setting, we find that the inclusion of the `no action' greatly benefits the localization setting, establishing a new state-of-the-art for Charades-localization (25.2 vs. 22.3 in \citep{piergiovanni2018tgm}).

\begin{table}
    \centering
    \caption{Fine-tuning on Charades using the currently available Kinetics videos. We report results for both classification and the localization setting. We also compare the use of the `none' action in AViD. [1] \citep{piergiovanni2018super}}
    \label{tab:finetune-charades}
    \begin{tabular}{c|cccc}
         \toprule
         Model & Pretrain Data & Class mAP & Loc mAP   \\
         \midrule
         I3D \citep{carreira2017quo} & Kin-400 & 34.3 & 17.9  \\
         I3D \citep{carreira2017quo} & Kin-600 & 36.5 & 18.4\\
         I3D \citep{carreira2017quo} & MiT & 33.5 & 15.4 \\
         I3D \citep{carreira2017quo} & AViD (- no action) & 36.2 & 17.3 \\
         I3D \citep{carreira2017quo} & AViD & 36.7 & 19.7 \\
        \midrule
         3D ResNet-50 & Kin-400 & 39.2 & 18.6  \\
         3D ResNet-50 & Kin-600 & 41.5 & 19.2 \\
         3D ResNet-50 & MiT & 35.4 & 16.4 \\  
         3D ResNet-50 & AViD (- no action) & 41.2 & 18.7 \\
         3D ResNet-50 & AViD & 41.7 & 23.2 \\
         3D ResNet-50 + super-events [1] & AViD & 42.4 & 25.2 \\
    \bottomrule
    \end{tabular}
\end{table}

\paragraph{Learning from Weak Tags}
We compare the effect of using the weak tags generated for the AViD dataset compared to using the manually labeled data. The results are shown in Table \ref{tab:cluster-hmdb}. Surprisingly, we find that using the weak tags provides strong initial features that can be fine-tuned on HMDB without much different in performance. Future works can explore how to best use the weak tag data. 

\begin{table}
    \centering
    \caption{Performance of 3D ResNet-50 using fully-labeled data vs. the weak tags data evaluated on HMDB. `Aff' is affinity propagation and `Agg' agglomerative clustering. }
    \label{tab:cluster-hmdb}
    \begin{tabular}{c|ccc}
         \toprule
         Model & Pretrain Data & Acc   \\
         \midrule
         3D ResNet-50 & Kin-400 & 76.7 \\
         3D ResNet-50 & AViD & 77.3 \\
         3D ResNet-50 & AViD-weak (Agg) & 76.4 \\
         3D ResNet-50 & AViD-weak (Aff) & 75.3 \\
    \bottomrule
    \end{tabular}
\end{table}

\paragraph{Blurred Face Effect}
During preprocessing, we use a face detector to blur any found faces in the videos. We utilize a strong Gaussian blur with random parameters. Gaussian blurring can be reversed if the location and parameters are known, however, due to the randomization of the parameters, it would be practically impossible to reverse the blur and recover true identity.

Since we are modifying the videos by blurring faces, we conducted experiments to see how face blurring impacts performance. We compare performance on AViD (accuracy) as well as fine-tuning on HMDB (accuracy) and Charades (mAP) classification. The results are shown in Table \ref{tab:face-blur}. While face blurring slightly reduces performance, the impact is not that great. This suggests it has a good balance of anonymization, yet still recognizable actions. 

\begin{table}
    \centering
    \caption{Measuring the effects of face blurring on AViD, HMDB and Charades classification. Note that only the faces in AViD are blurred.}
    \label{tab:face-blur}
    \begin{tabular}{c|cccc}
         \toprule
         Model & Data & AViD & HMDB & Charades   \\
         \midrule
         3D ResNet-50 & AViD-no blur & 48.2 & 77.5 & 42.1  \\
         3D ResNet-50 & AViD-blur & 47.9 & 77.3 & 41.7 \\
    \bottomrule
    \end{tabular}
    \vspace{-2mm}
\end{table}

\paragraph{Importance of Time}
In videos, the use of temporal information is often important when recognizing actions by using optical flow \citep{simonyan2014two}, stacking frames, RNNs \citep{ng2015beyond}, temporal pooling \citep{piergiovanni2017learning}, and other approaches. In order to determine how much temporal information AViD needs, we compared single-frame models to multi-frame. We then shuffled the frames to measure the performance drop. The results are shown in Table \ref{tab:temporal}. We find that adding more frames benefits performance, while shuffling them harms multi-frame model performance. This suggests that temporal information is quite useful for recognizing actions in AViD, making it an appropriate dataset for developing spatio-temporal video models.

\begin{table}
    \centering
    \caption{Effect of temporal information in AViD.}
    \label{tab:temporal}
    \begin{tabular}{c|ccc}
         \toprule
         Model & \# Frames & In Order & Shuffled  \\
         \midrule
         2D ResNet-50 & 1 & 32.5 & 32.5 \\
         3D ResNet-50 & 1 & 32.5 & 32.5 \\
         3D ResNet-50 & 16 & 44.5 & 38.7 \\
         3D ResNet-50 & 32 & 47.9 & 36.5 \\
         3D ResNet-50 & 64 & 48.2 & 35.6 \\
    \bottomrule
    \end{tabular}
\end{table}

\vspace{-1mm}
\section{Conclusions}
We present AViD, a new, static, diverse and anonymized video dataset. We showed the importance of collecting and learning from diverse videos, which is not captured in existing video datasets. Further, AViD is \textbf{static} and easily distributed, enabling reproducible research. Finally, we showed that AViD produces similar or better results on datasets like HMDB and Charades.

\section*{Broader Impacts}

We quantitatively confirmed that existing video datasets for action recognition are highly biased. In order to make people and researchers in diverse countries more fairly benefit from a public action recognition dataset, we propose the AViD dataset. We took care to query multiple websites from many countries in many languages to build a dataset that represents as many countries as possible. We experimentally showed that by doing this, we can reduce the bias of learned models.
We are not aware of any other large-scales datasets (with hundreds of video hours) which took such country diversity into the consideration during the collection process.





As this dataset contains a wide variety of actions, it could enable malicious parties to build systems to monitor people. However, we took many steps to preserve the identity of people and eliminate the ability to learn face-based actions, which greatly reduces the negative uses of the data. 
The positive impacts of this dataset are enabling reproducible research on video understanding which will help more advance video understanding research with consistent and reliable baselines. We emphasize once more that our dataset is a static dataset respecting the licences of all its videos.

\section*{Acknowledgement} This work was supported in part by the National Science Foundation (IIS-1812943 and CNS1814985).

{\small
\bibliography{bib}
}

\newpage
\clearpage

\appendix

\section{Diversity Statistics Collection}

In order to find the country location for each video in previous YouTube-based datasets (e.g., Kinetics, HACS, etc.), we used the public YouTube API.
Specifically, using \href{https://developers.google.com/youtube/v3/docs/videos}{https://developers.google.com/youtube/v3/docs/videos}, we extracted the `recordingDetails.location' object. Importantly, it notes that

\noindent \begin{quote}
    `The geolocation information associated with the video. Note that the child property values identify the location that the video owner wants to associate with the video. The value is editable, searchable on public videos, and might be displayed to users for public videos.'
\end{quote}
This is the only location data YouTube publicly provides and many videos in existing datasets do not have this field. In our measure, roughly 8\% of the videos had such geolocation. We then used \href{https://pypi.org/project/reverse-geocode/}{reverse-geocode library} \href{https://pypi.org/project/reverse-geocode/}{https://pypi.org/project/reverse-geocode/} to map the coordinates to the country, then manually mapped the countries to each region.

For full transparency, we provide detailed breakdowns of the diversity data we were able to measure with these tools in Table \ref{tab:my_label} as an example.

\begin{table}[hbt!]
    \centering
    \begin{tabular}{c|c}
    \toprule
        Country & Video Count \\
        \midrule
        North America & 32,767 \\
        EU & 1,613 \\
        Latin America & 2,289\\
        Asia & 938 \\
        Africa & 37 \\
        \midrule
        No Location & 422,645 \\
        \bottomrule
    \end{tabular}
    \caption{Kinetics-400 Video Distribution}
    \label{tab:my_label}
\end{table}

\section{Difference to Kinetics Numbers}
After the initial version of AViD was released (on arXiv), the Kinetics team provided numbers based on the estimated upload location of the video (this metadata is not publicly available) \citep{kineticsdiversity}.


In the paper, we have included their diversity statistics as well, as they are more complete, representing 90\% of videos, compared to about 8\% that we were able to get geolocation for.

{\small
\bibliography{bib}
}

\section{Action Classes}
\begin{multicols}{2}
\begin{enumerate}
    \item abseiling
\item acoustic guitar
\item acrobatic gymnastics
\item acting in play
\item adjusting glasses
\item aerobics
\item air drumming
\item air travel
\item airbrush
\item alligator wrestling
\item alpine climbing
\item alpine skiing
\item amusement park
\item answering questions
\item applauding
\item applying cream
\item archaeological excavation
\item archery
\item arguing
\item arm wrestling
\item arranging flowers
\item arresting
\item assembling bicycle
\item assembling computer
\item attending conference
\item auctioning
\item baby transport
\item baby waking up
\item backflip (human)
\item backpacking (wilderness)
\item baking
\item baking cookies
\item balance beam
\item balloon blowing
\item bandaging
\item barbell
\item barbequing
\item bartending
\item base jumping
\item bathing
\item bathing dog
\item batting (cricket)
\item batting cage
\item battle rope training
\item beatboxing
\item bee keeping
\item belly dancing
\item bench pressing
\item bending back
\item bending metal
\item biceps curl
\item bicycling
\item biking through snow
\item blasting sand
\item blending fruit
\item blowing glass
\item blowing leaves
\item blowing nose
\item blowing out candles
\item bmx bike
\item boating
\item bobsledding
\item body piercing
\item bodyboarding
\item bodysurfing
\item bodyweight exercise
\item bookbinding
\item bottling
\item bouncing ball
\item bouncing ball (not juggling)
\item bouncing on bouncy castle
\item bouncing on trampoline
\item bowling
\item bowling (cricket)
\item braiding hair
\item breading or breadcrumbing
\item breakdancing
\item breaking
\item breaking boards
\item breaking glass
\item breathing fire
\item brush painting
\item brushing hair
\item brushing teeth
\item building cabinet
\item building lego
\item building sandcastle
\item building shed
\item bull fighting
\item bulldozer
\item bulldozing
\item bungee jumping
\item burping
\item busking
\item buttoning
\item cake decorating
\item calculating
\item calligraphy
\item camping
\item canoeing or kayaking
\item capoeira
\item caporales
\item capsizing
\item card stacking
\item card throwing
\item card tricks
\item carp fishing
\item carrying baby
\item carrying weight
\item cartwheeling
\item carving ice
\item carving marble
\item carving pumpkin
\item carving wood with a knife
\item casting fishing line
\item catching fish
\item catching or throwing baseball
\item catching or throwing frisbee
\item catching or throwing softball
\item celebrating
\item changing gear in car
\item changing oil
\item changing wheel
\item chasing
\item checking tires
\item checking watch
\item cheerleading
\item chiseling stone
\item chiseling wood
\item chopping meat
\item chopping vegetables
\item chopping wood
\item christmas
\item circus
\item clam digging
\item clay pottery making
\item clean and jerk
\item cleaning floor
\item cleaning gutters
\item cleaning pool
\item cleaning shoes
\item cleaning toilet
\item cleaning windows
\item climbing a rope
\item climbing ladder
\item climbing tree
\item closing door
\item coloring in
\item combat
\item comedian
\item concert
\item construction
\item contact juggling
\item contorting
\item cooking
\item cooking chicken
\item cooking egg
\item cooking on campfire
\item cooking sausages
\item cooking sausages (not on barbeque)
\item cooking scallops
\item cooking show
\item cosplaying
\item counting money
\item country line dancing
\item cracking knuckles
\item cracking neck
\item crawling baby
\item cricket
\item crocheting
\item crossing river
\item crouching
\item crying
\item cumbia
\item curling (sport)
\item curling hair
\item cutting apple
\item cutting cake
\item cutting nails
\item cutting orange
\item cutting pineapple
\item cutting watermelon
\item dancing
\item dancing ballet
\item dancing charleston
\item dancing gangnam style
\item dancing macarena
\item dashcam
\item deadlifting
\item dealing cards
\item decorating the christmas tree
\item decoupage
\item delivering mail
\item demolition
\item digging
\item dining
\item directing traffic
\item dirt track racing
\item disc golfing
\item disc jockey
\item diving cliff
\item docking boat
\item dodgeball
\item dog agility
\item doing aerobics
\item doing jigsaw puzzle
\item doing laundry
\item doing nails
\item doing sudoku
\item doing wheelie
\item drag racing
\item drawing
\item dressage
\item dribbling basketball
\item drifting (motorsport)
\item drinking
\item drinking beer
\item drinking shots
\item driving car
\item driving tractor
\item drooling
\item drop kicking
\item drumming fingers
\item dumbbell
\item dump truck
\item dumpster diving
\item dune buggy
\item dunking basketball
\item dying hair
\item eating burger
\item eating cake
\item eating carrots
\item eating chips
\item eating doughnuts
\item eating hotdog
\item eating ice cream
\item eating nachos
\item eating spaghetti
\item eating street food
\item eating watermelon
\item egg hunting
\item electric guitar
\item embroidering
\item embroidery
\item enduro
\item entering church
\item exercising arm
\item exercising with an exercise ball
\item explosion
\item extinguishing fire
\item extreme sport
\item faceplanting
\item falling off bike
\item falling off chair
\item feeding birds
\item feeding fish
\item feeding goats
\item building fence
\item fencing (sport)
\item festival
\item fidgeting
\item field hockey
\item figure skating
\item filling cake
\item filling eyebrows
\item finger snapping
\item fingerboard (skateboard)
\item firefighter
\item fireworks
\item fixing bicycle
\item fixing hair
\item flamenco
\item flint knapping
\item flipping bottle
\item flipping pancake
\item fly tying
\item flying kite
\item folding clothes
\item folding napkins
\item folding paper
\item forklift
\item french horn
\item front raises
\item frying
\item frying vegetables
\item gambling
\item garbage collecting
\item gardening
\item gargling
\item geocaching
\item getting a haircut
\item getting a piercing
\item getting a tattoo
\item giving or receiving award
\item gliding
\item go-kart
\item gold panning
\item golf chipping
\item golf driving
\item golf putting
\item gospel singing in church
\item greeting
\item grinding meat
\item grooming cat
\item grooming dog
\item grooming horse
\item gymnastics
\item gymnastics tumbling
\item hammer throw
\item hand washing clothes
\item head stand
\item headbanging
\item headbutting
\item heavy equipment
\item helmet diving
\item herding cattle
\item high fiving
\item high jump
\item high kick
\item hiking
\item historical reenactment
\item hitchhiking
\item hitting baseball
\item hockey stop
\item holding snake
\item home improvement
\item home roasting coffee
\item hopscotch
\item horse racing
\item hoverboarding
\item huddling
\item hugging
\item hugging (not baby)
\item hugging baby
\item hula hooping
\item hunting
\item hurdling
\item hurling (sport)
\item ice climbing
\item ice dancing
\item ice fishing
\item ice skating
\item ice swimming
\item inflating balloons
\item installing carpet
\item ironing
\item ironing hair
\item javelin throw
\item jaywalking
\item jetskiing
\item jogging
\item juggling
\item juggling balls
\item juggling fire
\item juggling soccer ball
\item jumping
\item jumping bicycle
\item jumping into pool
\item jumping jacks
\item jumping sofa
\item jumpstyle dancing
\item karaoke
\item kick (football)
\item kickboxing
\item kickflip
\item kicking field goal
\item kicking soccer ball
\item kissing
\item kitesurfing
\item knitting
\item krumping
\item land sailing
\item landing airplane
\item laughing
\item lawn mower racing
\item laying bricks
\item laying concrete
\item laying decking
\item laying stone
\item laying tiles
\item leatherworking
\item letting go of balloon
\item licking
\item lifting hat
\item lighting
\item lighting candle
\item lighting fire
\item listening with headphones
\item lock picking
\item logging
\item long jump
\item longboarding
\item looking at phone
\item looking in mirror
\item luge
\item lunge
\item making a cake
\item making a sandwich
\item making balloon shapes
\item making bed
\item making bubbles
\item making cheese
\item making horseshoes
\item making jewelry
\item making latte art
\item making paper aeroplanes
\item making pizza
\item making snowman
\item making sushi
\item making tea
\item making the bed
\item manicure
\item manufacturing
\item marching
\item marching band
\item marimba
\item marriage proposal
\item massaging back
\item massaging feet
\item massaging legs
\item massaging neck
\item mechanic
\item metal detecting
\item metal working
\item milking cow
\item milking goat
\item minibike
\item mixing colours
\item model building
\item monster truck
\item moon walking
\item mopping floor
\item mosh pit dancing
\item motocross
\item motorcycling
\item mountain biking
\item mountain climber (exercise)
\item moving baby
\item moving child
\item moving furniture
\item mowing lawn
\item mushroom foraging
\item musical ensemble
\item needle felting
\item news anchoring
\item news presenter
\item nightclub
\item none
\item offroading
\item ollie (skateboarding)
\item omelette
\item opening bottle
\item opening bottle (not wine)
\item opening coconuts
\item opening door
\item opening present
\item opening refrigerator
\item opening wine bottle
\item orchestra
\item origami
\item outdoor recreation
\item packing
\item parade
\item paragliding
\item parasailing
\item parkour
\item passing american football
\item passing soccer ball
\item peeling apples
\item peeling banana
\item peeling potatoes
\item penalty kick (association football)
\item person collecting garbage
\item personal computer
\item petting animal
\item petting animal (not cat)
\item petting cat
\item petting horse
\item photobombing
\item photocopying
\item picking apples
\item picking blueberries
\item picking fruit
\item pilates
\item pillow fight
\item pinching
\item pipe organ
\item pirouetting
\item planing wood
\item planting trees
\item plastering
\item playing accordion
\item playing american football
\item playing badminton
\item playing bagpipes
\item playing banjo
\item playing basketball
\item playing bass guitar
\item playing beer pong
\item playing billiards
\item playing blackjack
\item playing cards
\item playing cello
\item playing checkers
\item playing chess
\item playing clarinet
\item playing controller
\item playing cricket
\item playing cymbals
\item playing darts
\item playing didgeridoo
\item playing dominoes
\item playing drums
\item playing fiddle
\item playing field hockey
\item playing flute
\item playing gong
\item playing guitar
\item playing hand clapping games
\item playing handball
\item playing harmonica
\item playing harp
\item playing ice hockey
\item playing keyboard
\item playing kickball
\item playing laser tag
\item playing lute
\item playing mahjong
\item playing maracas
\item playing marbles
\item playing monopoly
\item playing netball
\item playing oboe
\item playing ocarina
\item playing organ
\item playing paintball
\item playing pan pipes
\item playing piano
\item playing piccolo
\item playing pinball
\item playing ping pong
\item playing poker
\item playing polo
\item playing recorder
\item playing road hockey
\item playing rounders
\item playing rubiks cube
\item playing rugby
\item playing saxophone
\item playing scrabble
\item playing shuffleboard
\item playing slot machine
\item playing snare drum
\item playing soccer
\item playing squash or racquetball
\item playing tennis
\item playing timbales
\item playing trombone
\item playing trumpet
\item playing tuba
\item playing ukulele
\item playing viola
\item playing violin
\item playing volleyball
\item playing with toys
\item playing with trains
\item playing xylophone
\item plumbing
\item poaching eggs
\item poking bellybutton
\item pole vault
\item polishing furniture
\item polishing metal
\item popping balloons
\item pouring beer
\item pouring milk
\item pouring wine
\item praying
\item preacher
\item preparing salad
\item presenting weather forecast
\item pretending to be a statue
\item protesting
\item pull ups
\item pulling
\item pulling espresso shot
\item pulling rope
\item pulling rope (game)
\item pumping fist
\item pumping gas
\item punching bag
\item punching person
\item push up
\item pushing car
\item pushing cart
\item pushing wheelbarrow
\item pushing wheelchair
\item putting on foundation
\item putting on lipstick
\item putting on sari
\item putting on shoes
\item putting wallpaper on wall
\item queuing
\item racing
\item radio-controlled model
\item rafting
\item rain
\item rallying
\item reading book
\item reading newspaper
\item recipe
\item recording music
\item recreational fishing
\item repairing puncture
\item riding a bike
\item riding camel
\item riding elephant
\item riding mechanical bull
\item riding mule
\item riding or walking with horse
\item riding scooter
\item riding snow blower
\item riding unicycle
\item ripping paper
\item roasting
\item roasting marshmallows
\item roasting pig
\item robot dancing
\item rock climbing
\item rock scissors paper
\item rocking
\item roller coaster
\item roller skating
\item rolling pastry
\item rope pushdown
\item rowing (sport)
\item running
\item running on treadmill
\item sailing
\item salsa dancing
\item saluting
\item sanding floor
\item sanding wood
\item sausage making
\item sawing wood
\item scrambling eggs
\item scrapbooking
\item screen printing
\item scrubbing face
\item scuba diving
\item seasoning food
\item separating eggs
\item serve (tennis)
\item setting table
\item sewing
\item shaking hands
\item shaking head
\item shaping bread dough
\item sharpening knives
\item sharpening pencil
\item shaving head
\item shaving legs
\item shearing sheep
\item shining flashlight
\item shining shoes
\item shooting basketball
\item shooting off fireworks
\item shopping
\item shot put
\item shouting
\item shoveling snow
\item shredding paper
\item shrugging
\item shucking oysters
\item shuffling cards
\item shuffling feet
\item side kick
\item sieving
\item sign language interpreting
\item silent disco
\item singing
\item sipping cup
\item situp
\item skateboarding
\item ski ballet
\item ski jumping
\item skiing crosscountry
\item skiing mono
\item skiing slalom
\item skipping rope
\item skipping stone
\item sky diving
\item skydiving
\item slacklining
\item slapping
\item sled dog racing
\item sleeping
\item slicing onion
\item slopestyle
\item smashing
\item smelling feet
\item smoking
\item smoking hookah
\item smoking pipe
\item smoothie
\item snatch weight lifting
\item sneezing
\item sniffing
\item snorkeling
\item snowboarding
\item snowkiting
\item snowmobile
\item snowmobiling
\item snowplow
\item snowshoe
\item soccer goal
\item somersaulting
\item sowing
\item speed skating
\item spelunking
\item spinning plates
\item spinning poi
\item splashing
\item splashing water
\item spray painting
\item spraying
\item springboard diving
\item square dancing
\item squat
\item squeezing orange
\item stacking cups
\item stacking dice
\item standing on hands
\item standup paddleboarding
\item staring
\item stealing
\item steer roping
\item steering car
\item sticking tongue out
\item stir frying
\item stirring
\item stomping grapes
\item street racing
\item stretching
\item stretching arm
\item stretching leg
\item strumming guitar
\item stunt performer
\item submerging
\item sucking lolly
\item sun tanning
\item surfing crowd
\item surfing water
\item surveying
\item sweeping floor
\item swimming
\item swimming backstroke
\item swimming breast stroke
\item swimming butterfly stroke
\item swimming front crawl
\item swimming with dolphins
\item swimming with sharks
\item swing dancing
\item swinging baseball bat
\item swinging legs
\item swinging on something
\item sword fighting
\item sword swallowing
\item tabla
\item tackling
\item tagging graffiti
\item tai chi
\item taking a shower
\item taking photo
\item talking on cell phone
\item tango dancing
\item tap dancing
\item tapping guitar
\item tapping pen
\item tasting beer
\item tasting food
\item tasting wine
\item teaching
\item tearing
\item telemark ski
\item tennis
\item testifying
\item texting
\item threading needle
\item throwing axe
\item throwing ball
\item throwing ball (not baseball or american football)
\item throwing discus
\item throwing knife
\item throwing snowballs
\item throwing tantrum
\item throwing water balloon
\item thunderstorm
\item tickling
\item tie dying
\item tightrope walking
\item tiptoeing
\item tobogganing
\item torte
\item tossing coin
\item tossing salad
\item train
\item training dog
\item trapezing
\item treating wood
\item trimming or shaving beard
\item trimming shrubs
\item trimming trees
\item triple jump
\item twiddling fingers
\item tying bow tie
\item tying knot (not on a tie)
\item tying necktie
\item tying shoe laces
\item tying tie
\item unboxing
\item uncorking champagne
\item underwater diving
\item unidentified flying object
\item unloading truck
\item using a microscope
\item using a paint roller
\item using a power drill
\item using a sledge hammer
\item using a wrench
\item using atm
\item using bagging machine
\item using circular saw
\item using computer
\item using inhaler
\item using megaphone
\item using puppets
\item using remote controller
\item using remote controller (not gaming)
\item using segway
\item vacuum cleaner
\item vacuuming car
\item vacuuming floor
\item valuting
\item visiting the zoo
\item volcano
\item wading through mud
\item wading through water
\item waiting in line
\item wakeboarding
\item waking up
\item walking on stilts
\item walking the dog
\item walking through snow
\item walking with crutches
\item washing
\item washing dishes
\item washing feet
\item washing hair
\item washing hands
\item washing machine
\item watching tv
\item water park
\item water skiing
\item water sliding
\item watercolor painting
\item waterfall
\item waterfowl hunting
\item watering plants
\item waving hand
\item waxing armpits
\item waxing back
\item waxing chest
\item waxing eyebrows
\item waxing legs
\item weaving basket
\item weaving fabric
\item wedding
\item weight lifting
\item welding
\item whistling
\item wildlife
\item windsurfing
\item winking
\item wood burning (art)
\item wood carving
\item woodworking
\item wrapping present
\item wrestling
\item writing
\item yarn spinning
\item yawning
\item yoga
\item zumba

\end{enumerate}
\end{multicols}

\section{Full Hierarchy}
\begin{figure}
    \centering
    \includegraphics[width=0.9\linewidth]{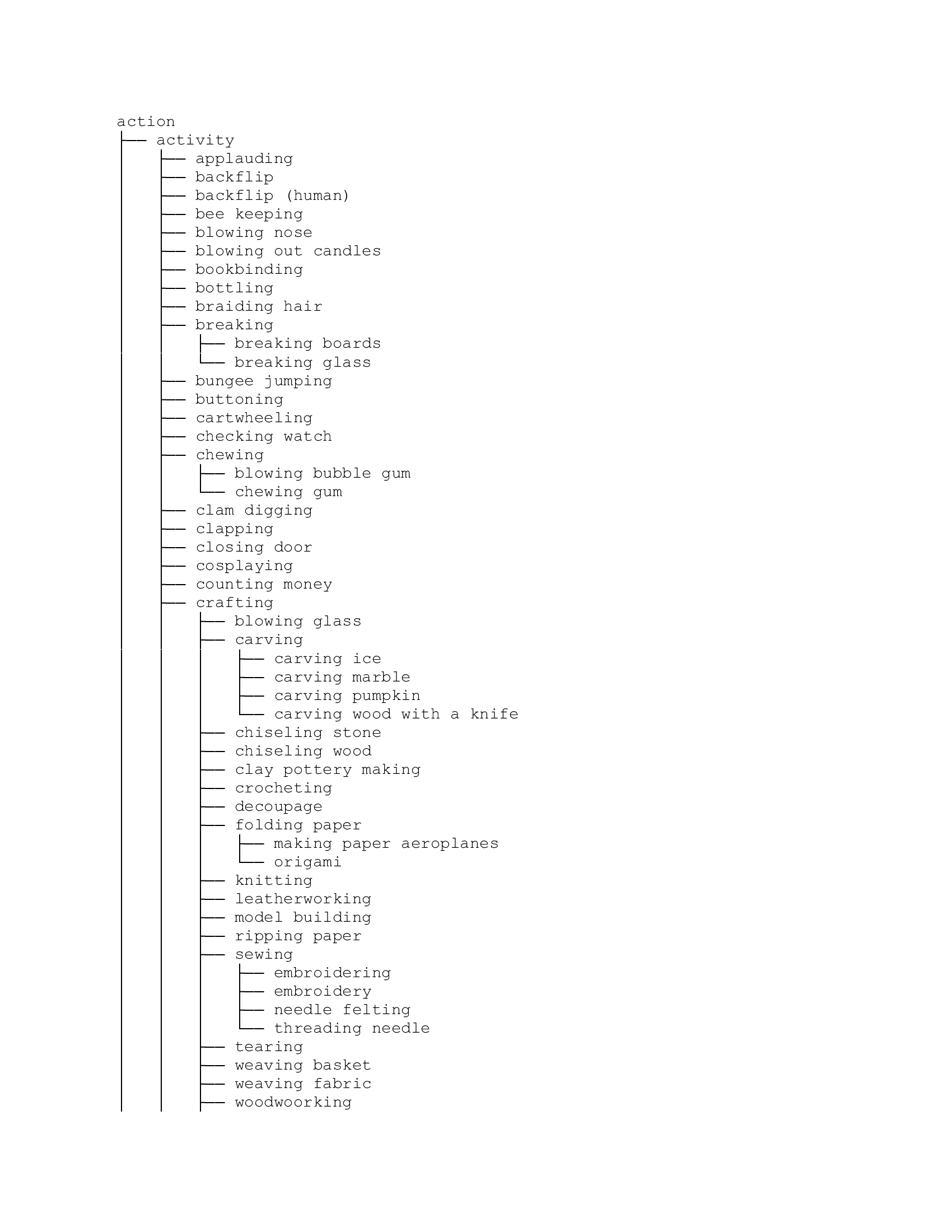}
        \end{figure}
        \begin{figure}
        \centering
    \includegraphics[width=0.9\linewidth]{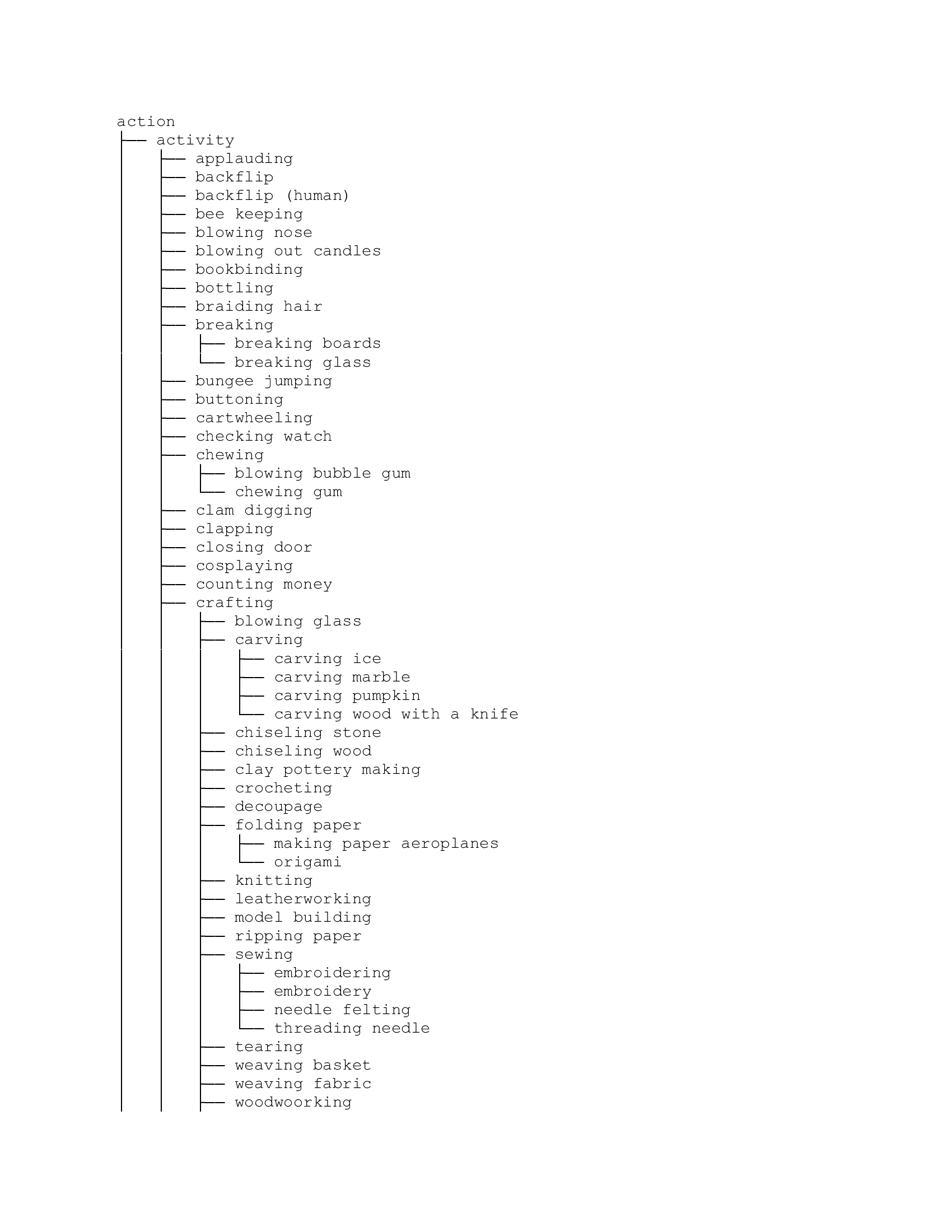}      \end{figure}
        \begin{figure}
        \centering
    \includegraphics[width=0.9\linewidth]{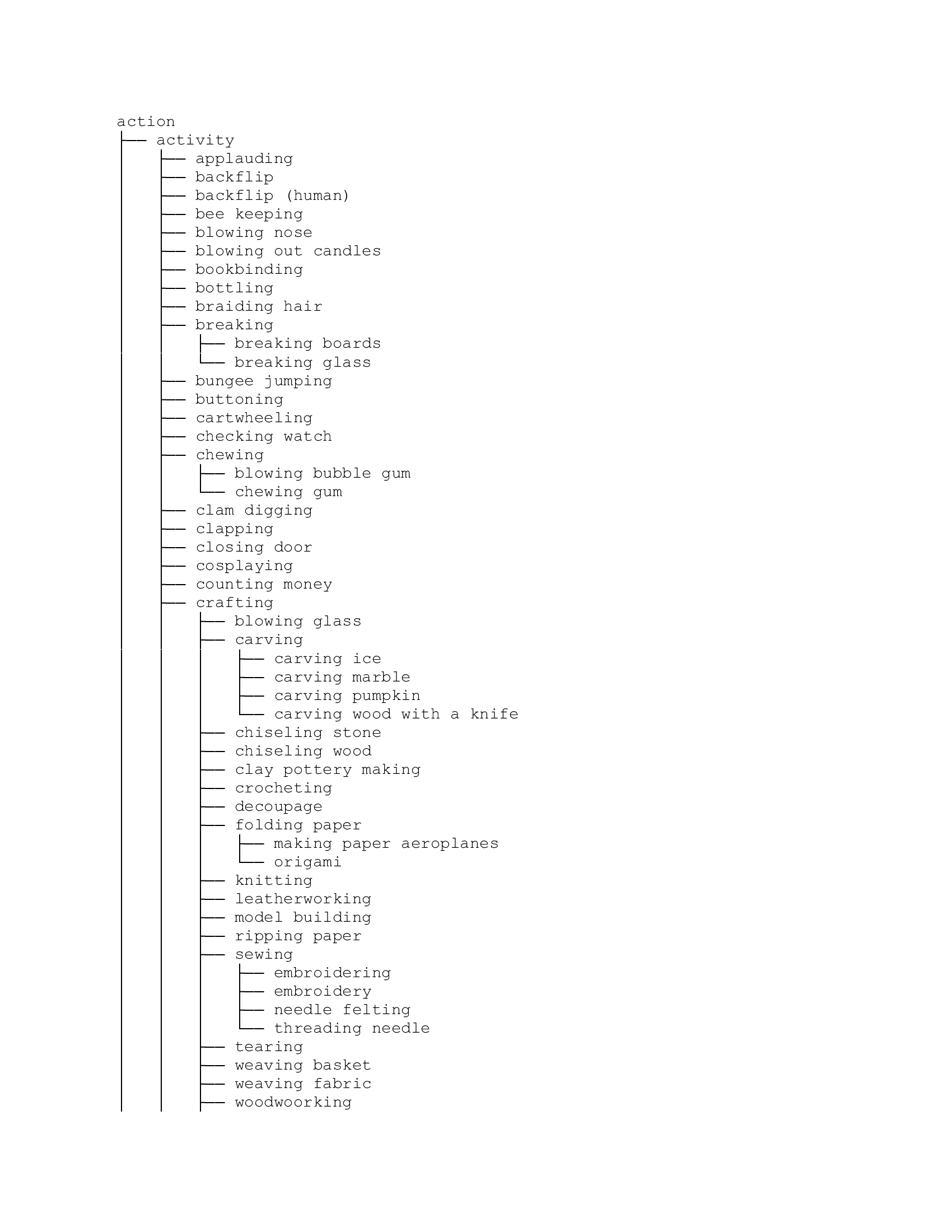}      \end{figure}
        \begin{figure}
        \centering
    \includegraphics[width=0.9\linewidth]{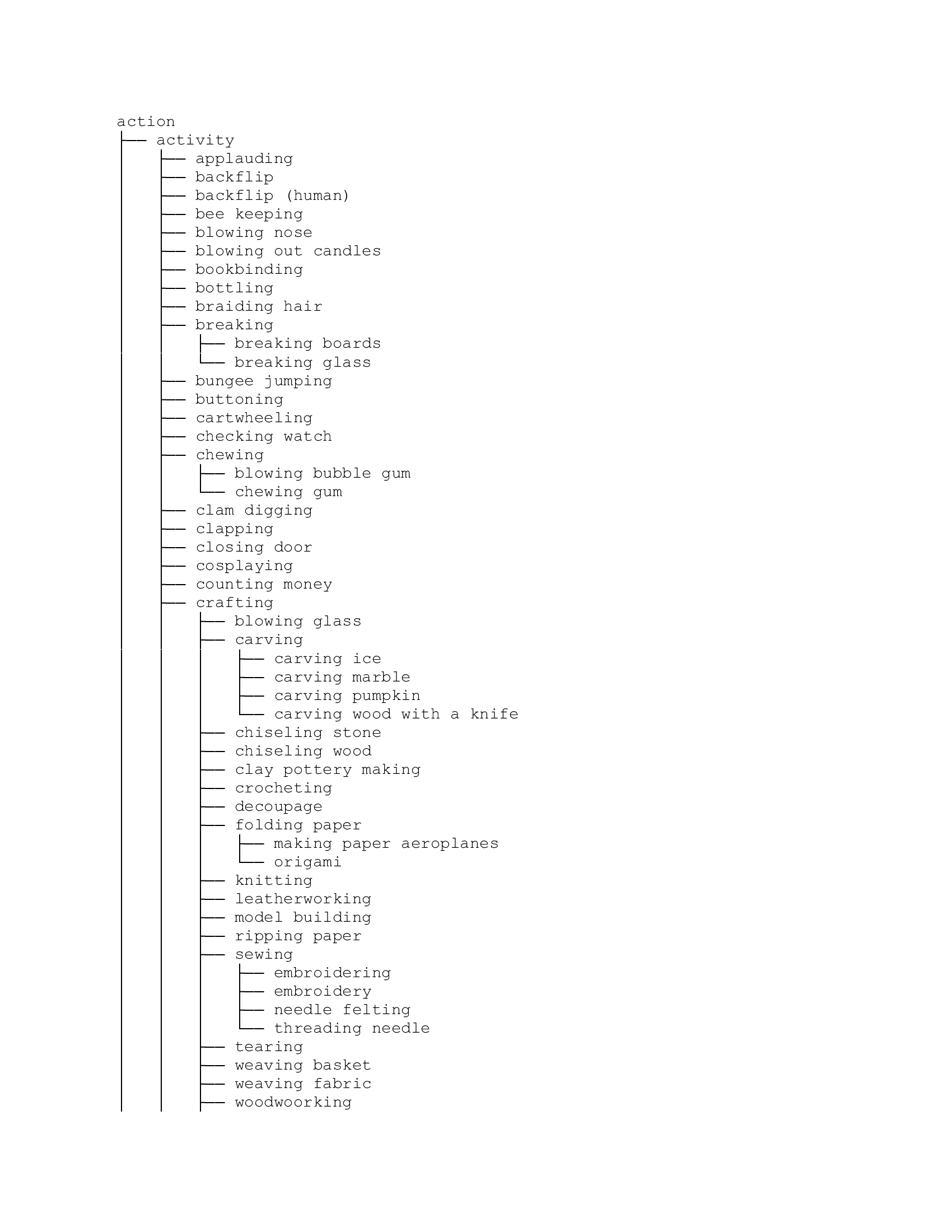}      \end{figure}
        \begin{figure}
        \centering
    \includegraphics[width=0.9\linewidth]{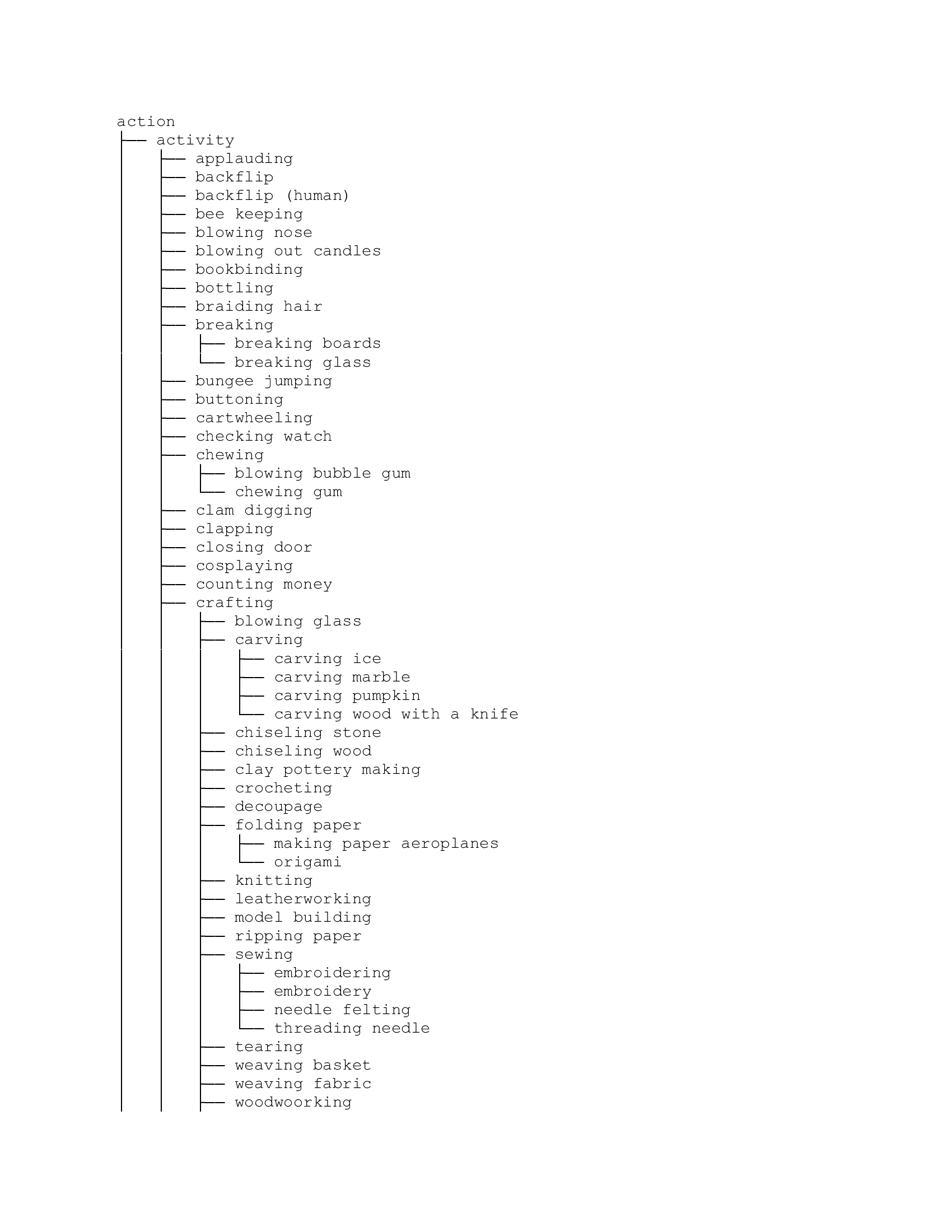}      \end{figure}
        \begin{figure}
        \centering
    \includegraphics[width=0.9\linewidth]{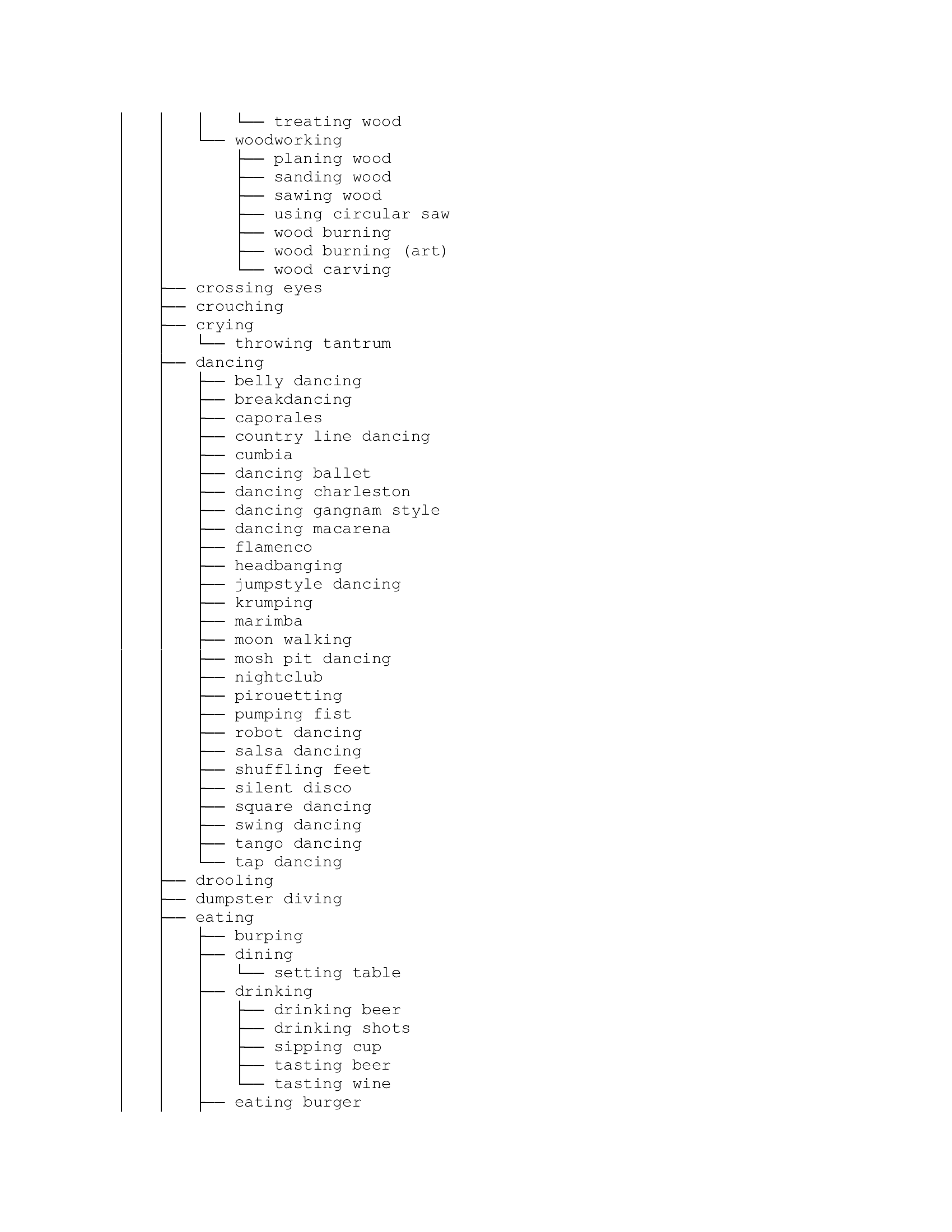}      \end{figure}
        \begin{figure}
        \centering
    \includegraphics[width=0.9\linewidth]{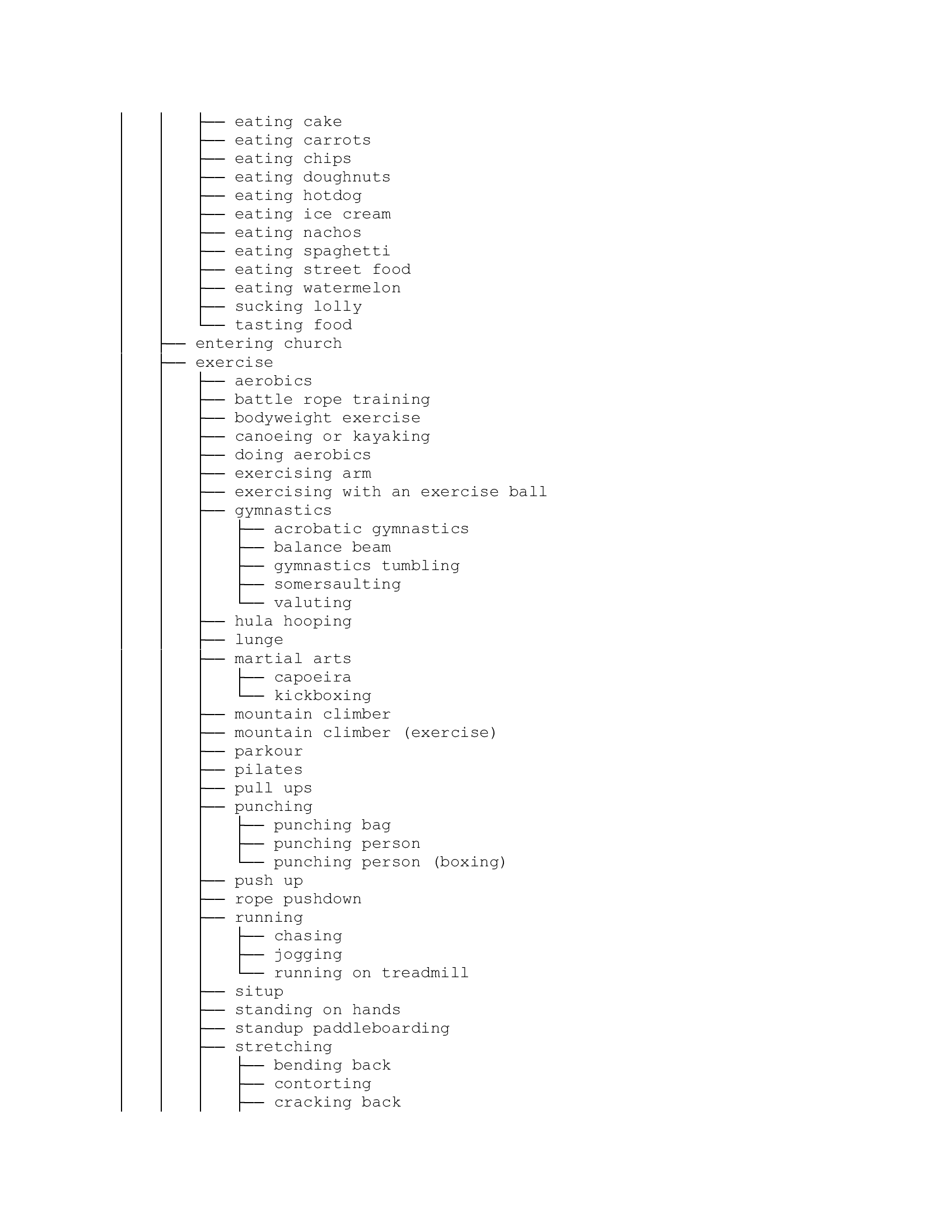}      \end{figure}
        \begin{figure}
        \centering
    \includegraphics[width=0.9\linewidth]{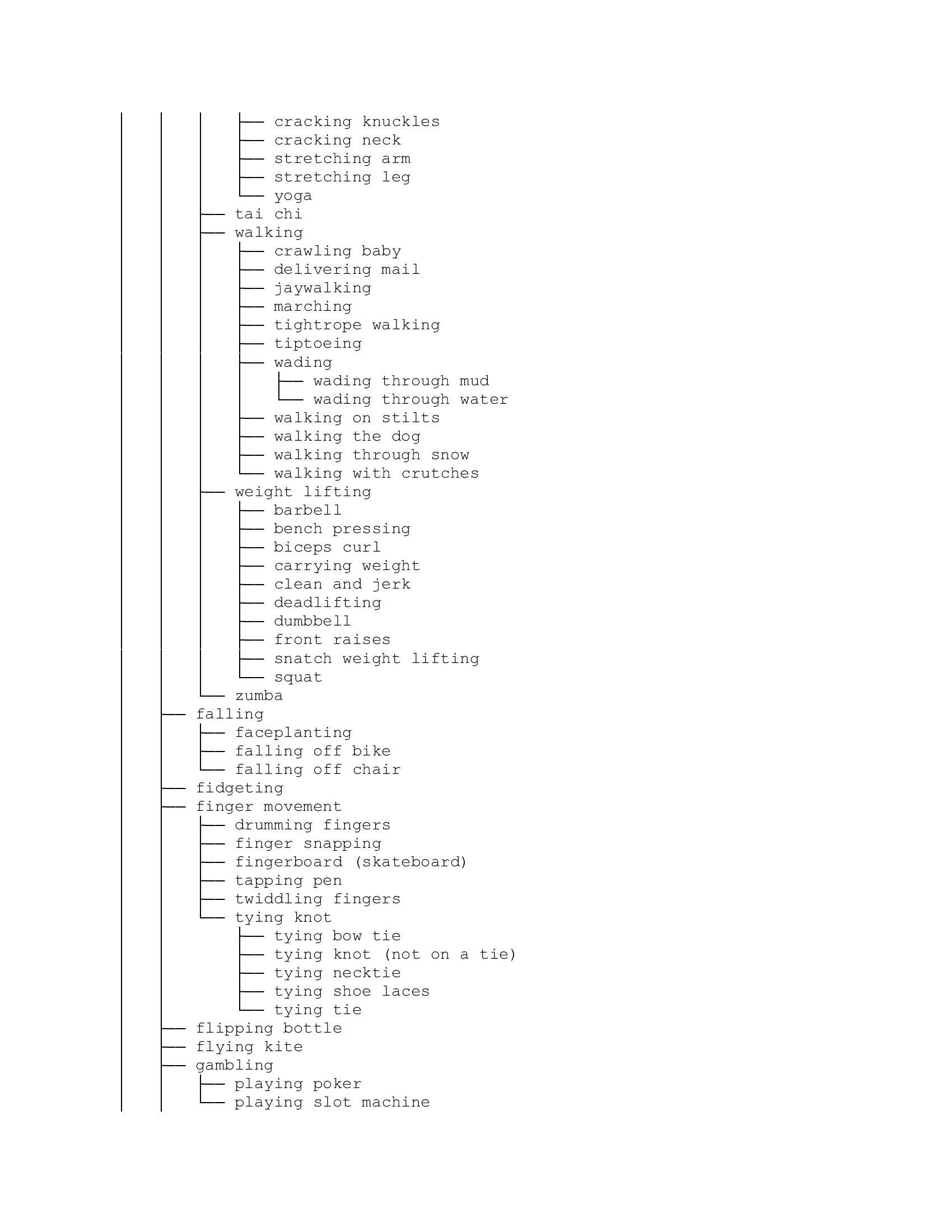}      \end{figure}
        \begin{figure}
        \centering
    \includegraphics[width=0.9\linewidth]{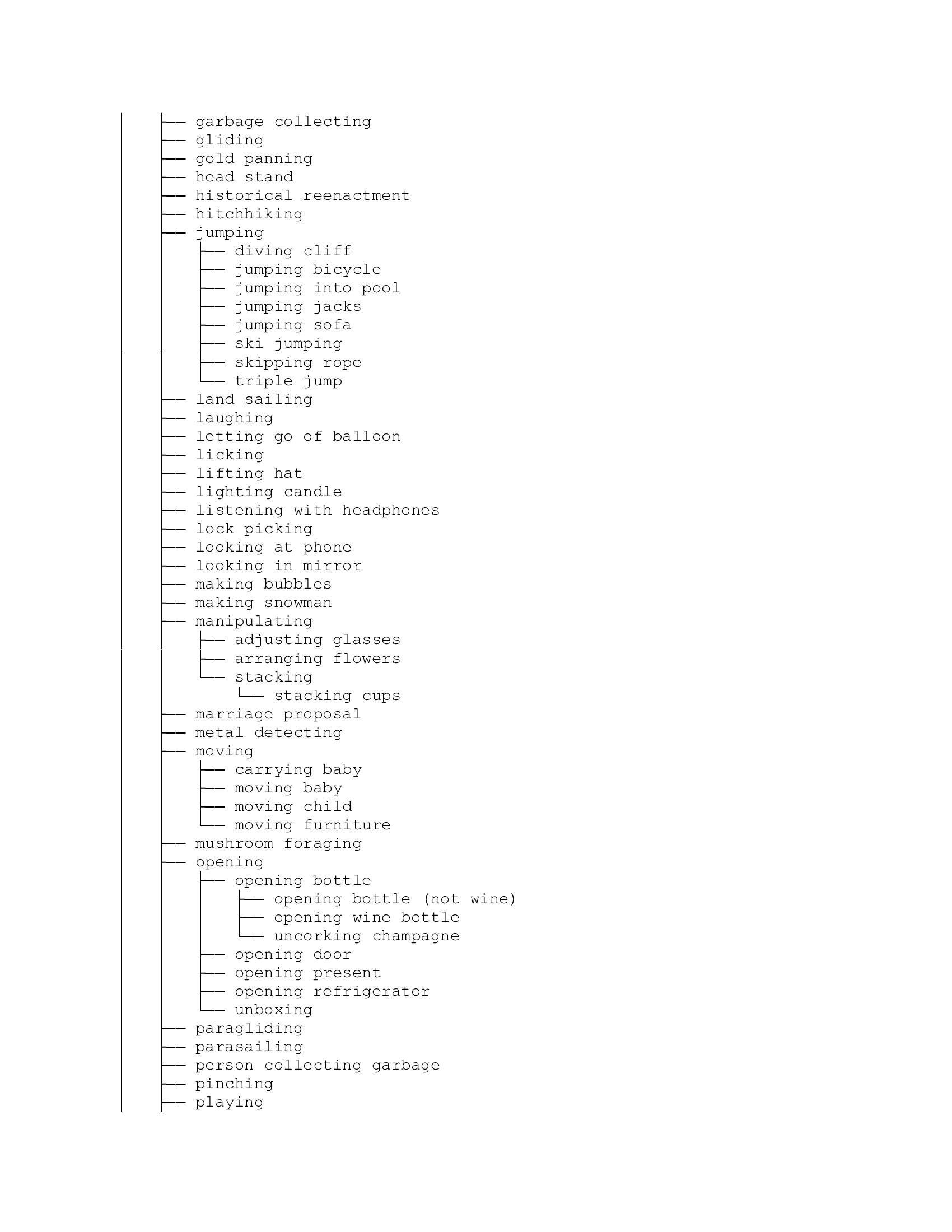}      \end{figure}
        \begin{figure}
        \centering
    \includegraphics[width=0.9\linewidth]{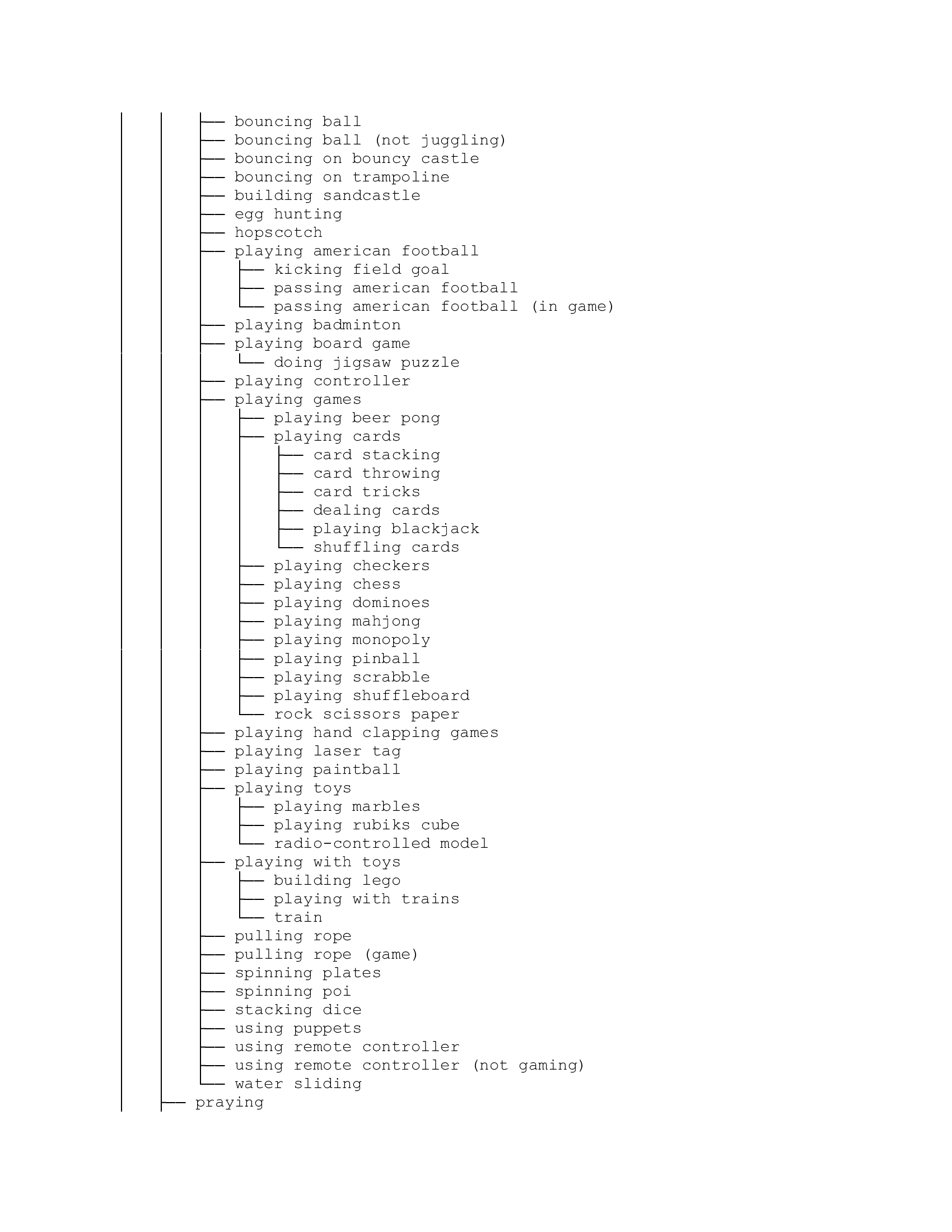}      \end{figure}
        \begin{figure}
        \centering
    \includegraphics[width=0.9\linewidth]{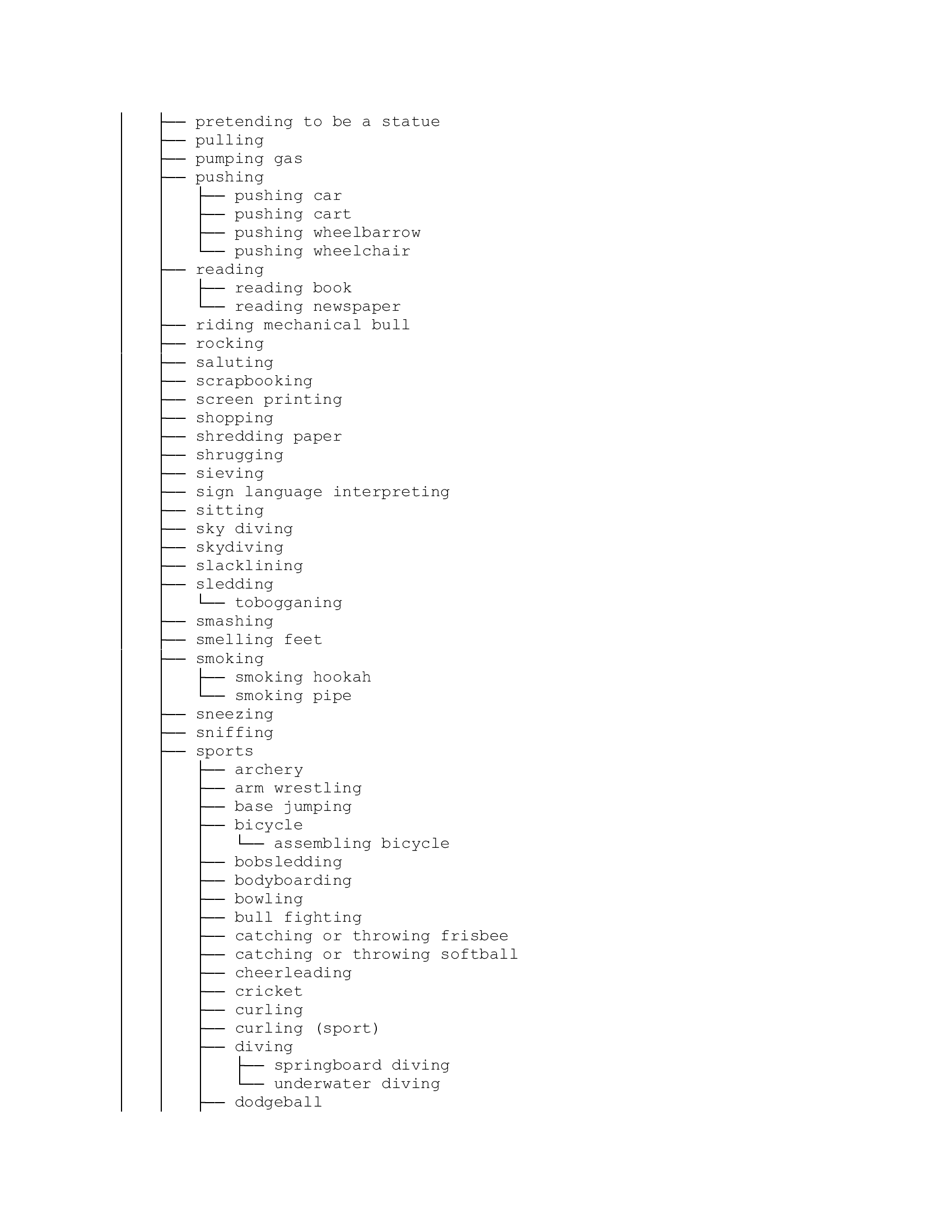}      \end{figure}
        \begin{figure}
        \centering
    \includegraphics[width=0.9\linewidth]{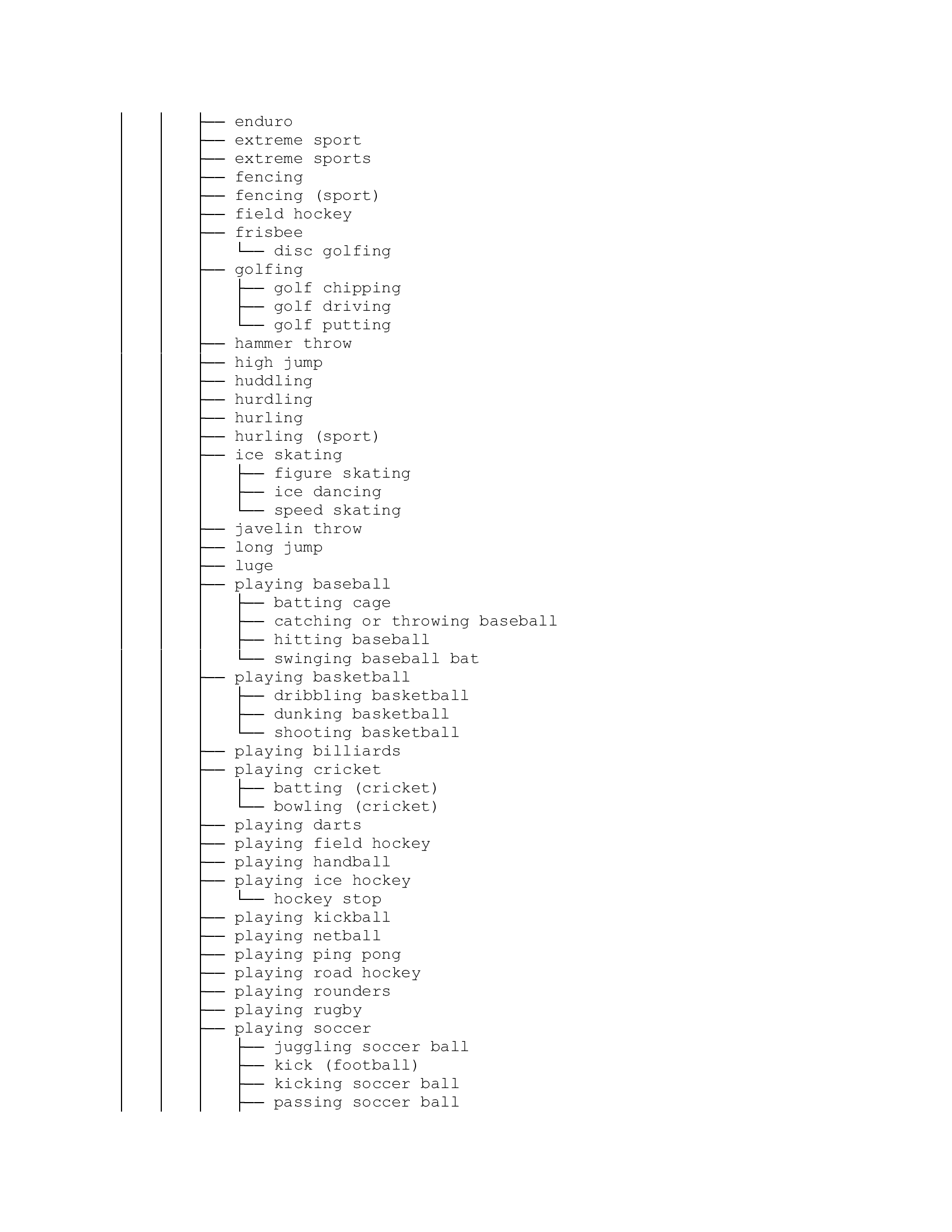}      \end{figure}
        \begin{figure}
        \centering
    \includegraphics[width=0.9\linewidth]{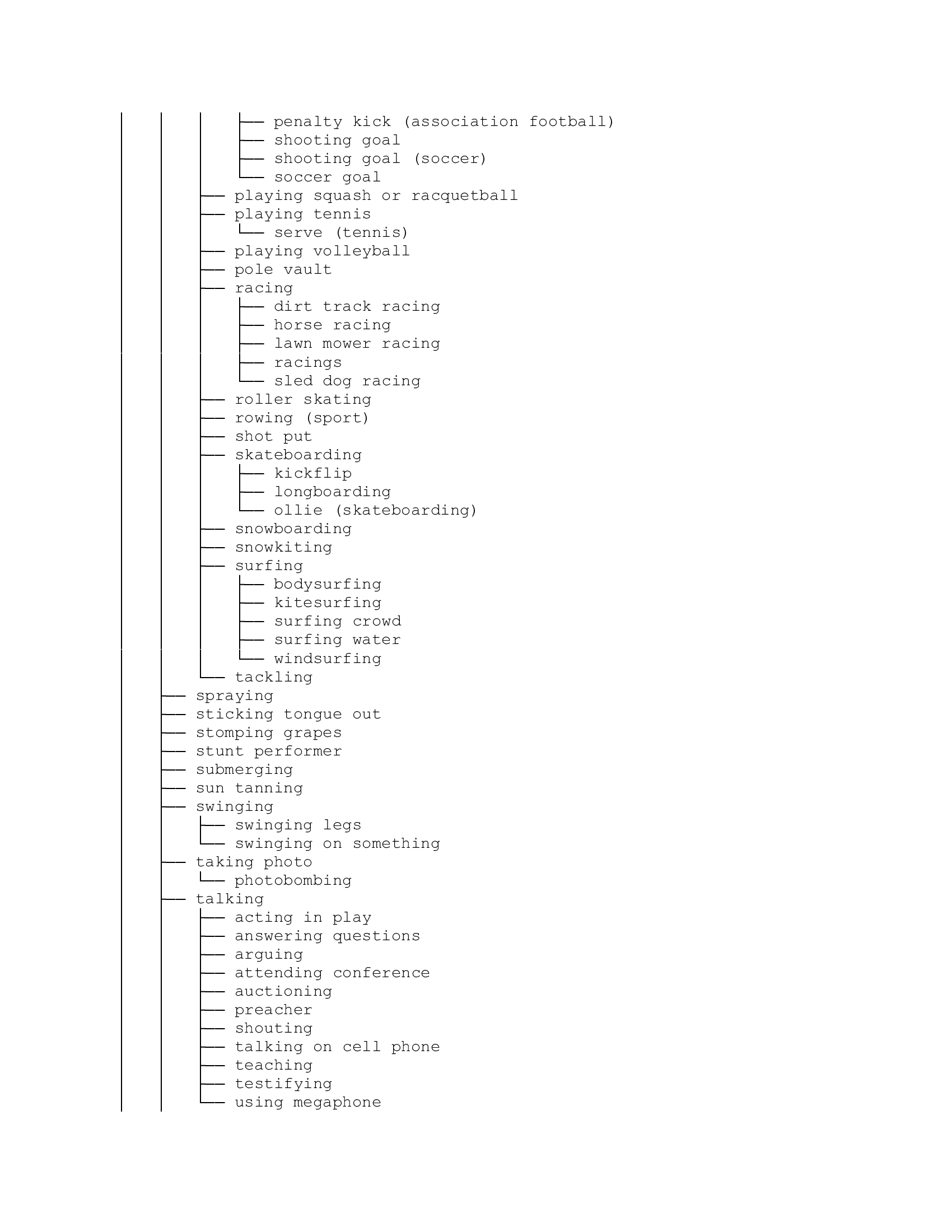}      \end{figure}
        \begin{figure}
        \centering
    \includegraphics[width=0.9\linewidth]{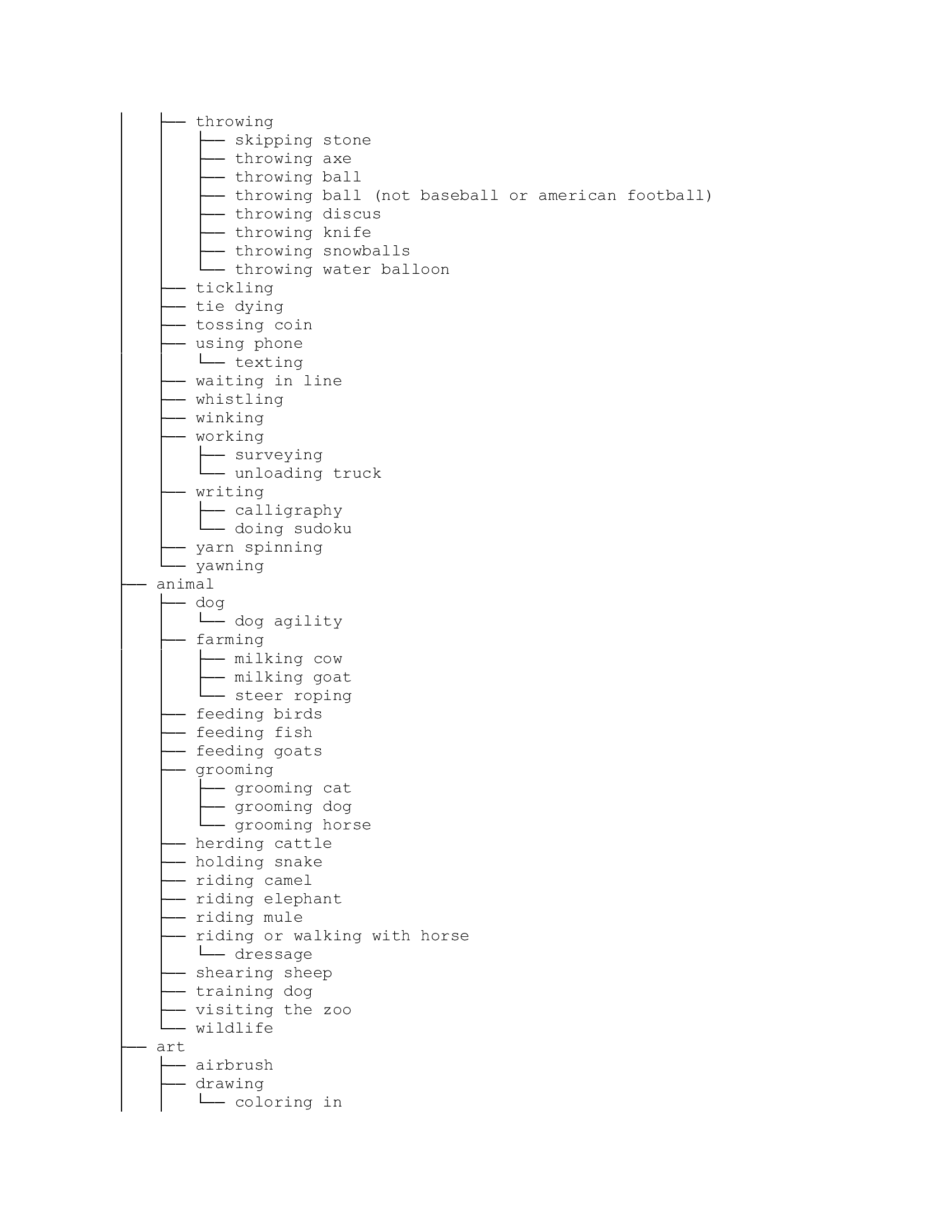}      \end{figure}
        \begin{figure}
        \centering
    \includegraphics[width=0.9\linewidth]{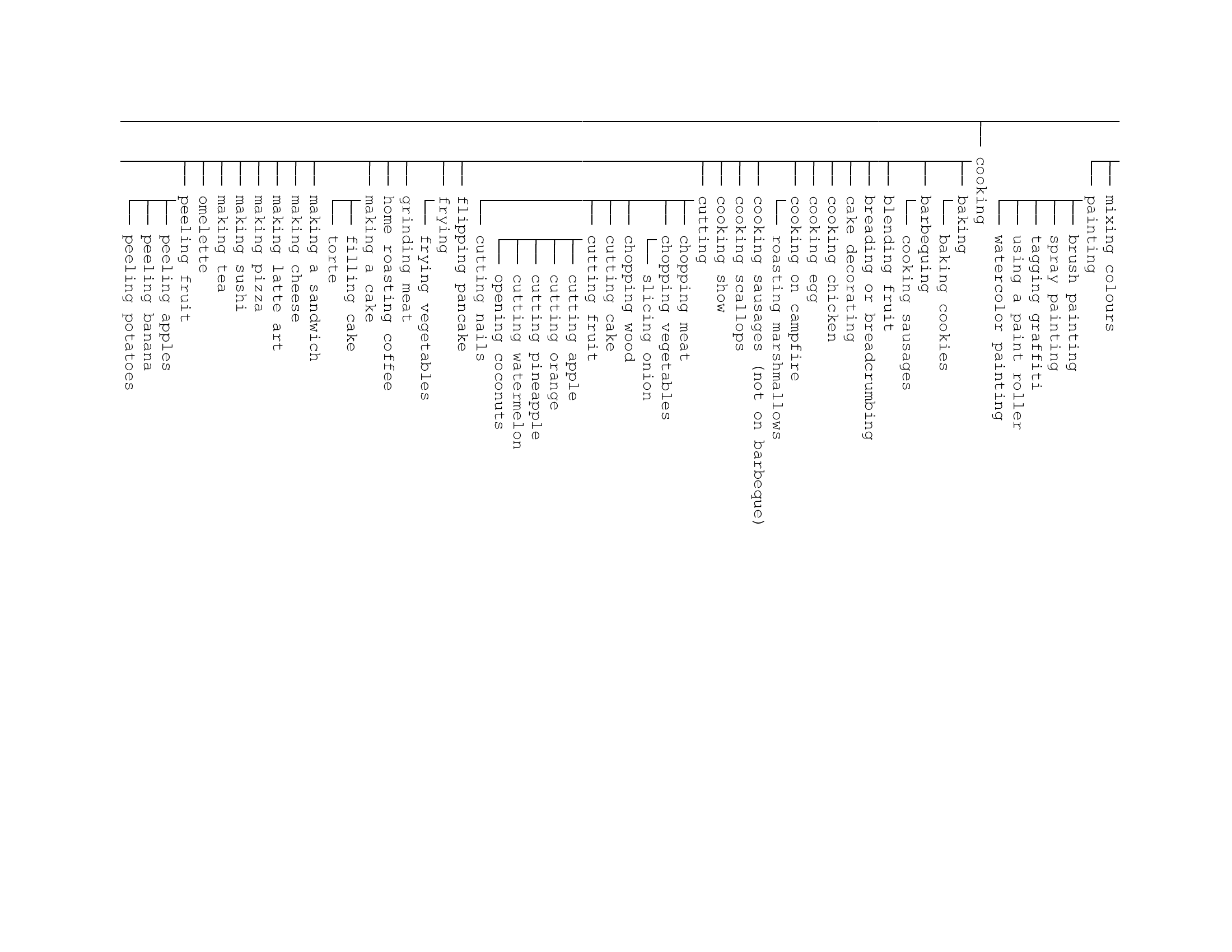}      \end{figure}
        \begin{figure}
        \centering
    \includegraphics[width=0.9\linewidth]{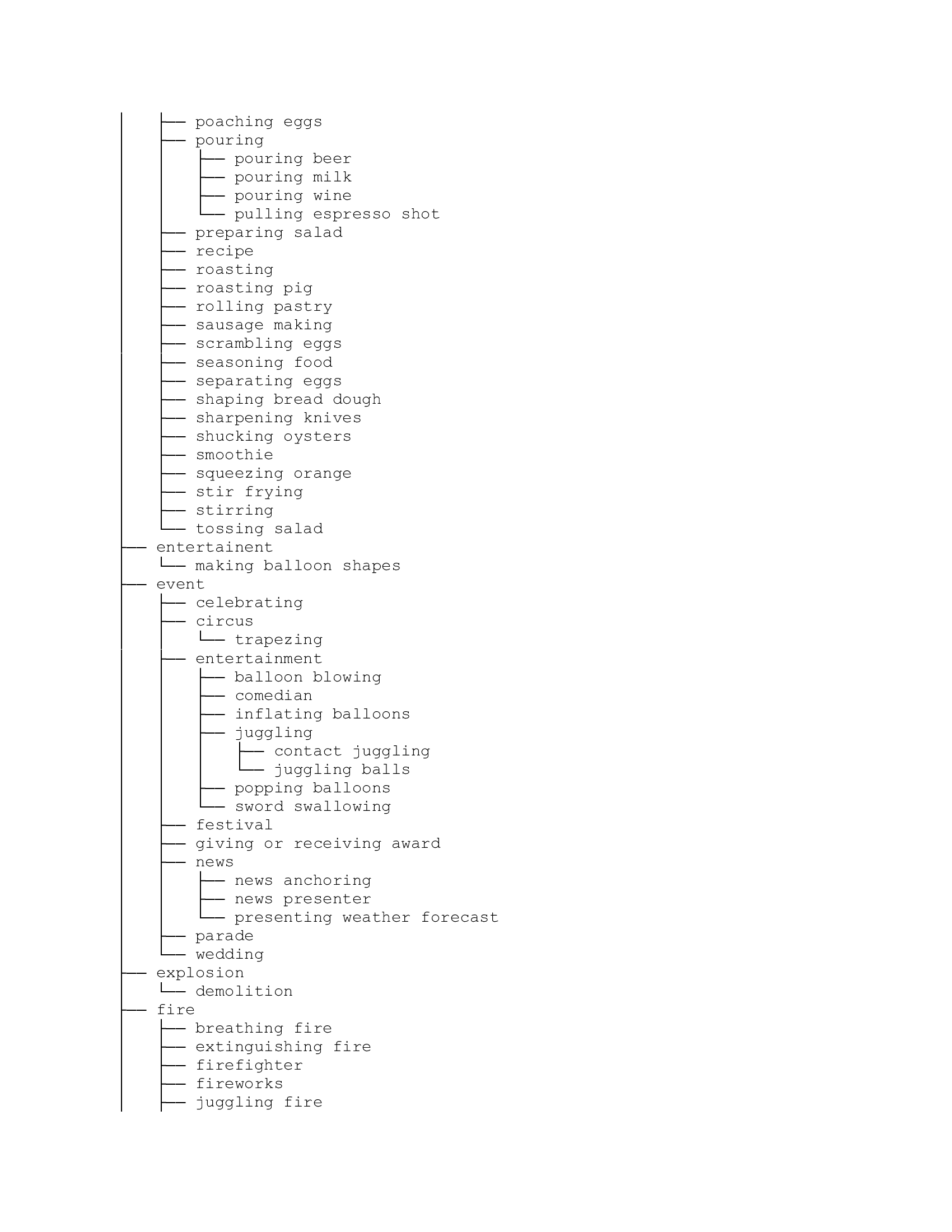}      \end{figure}
        \begin{figure}
        \centering
    \includegraphics[width=0.9\linewidth]{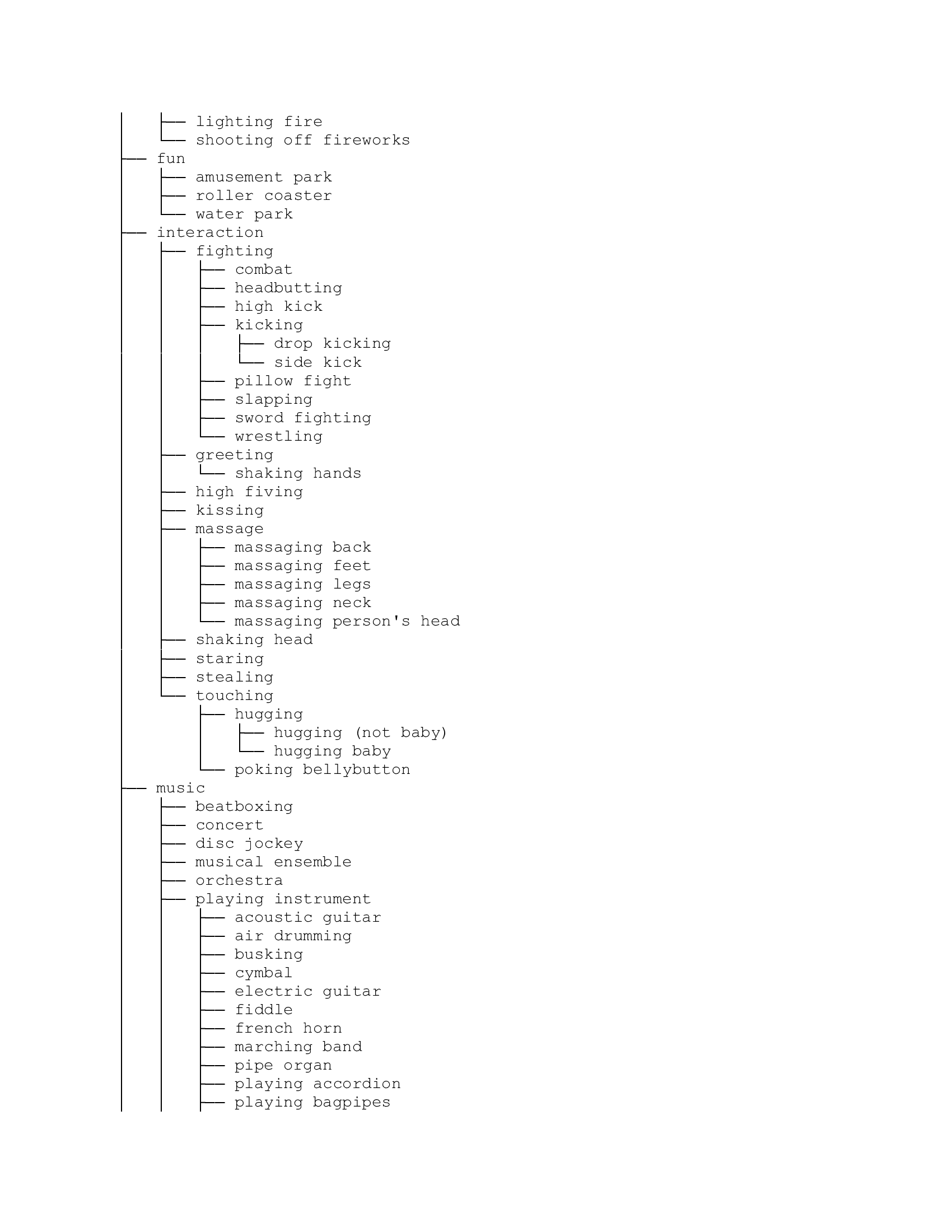}      \end{figure}
        \begin{figure}
        \centering
    \includegraphics[width=0.9\linewidth]{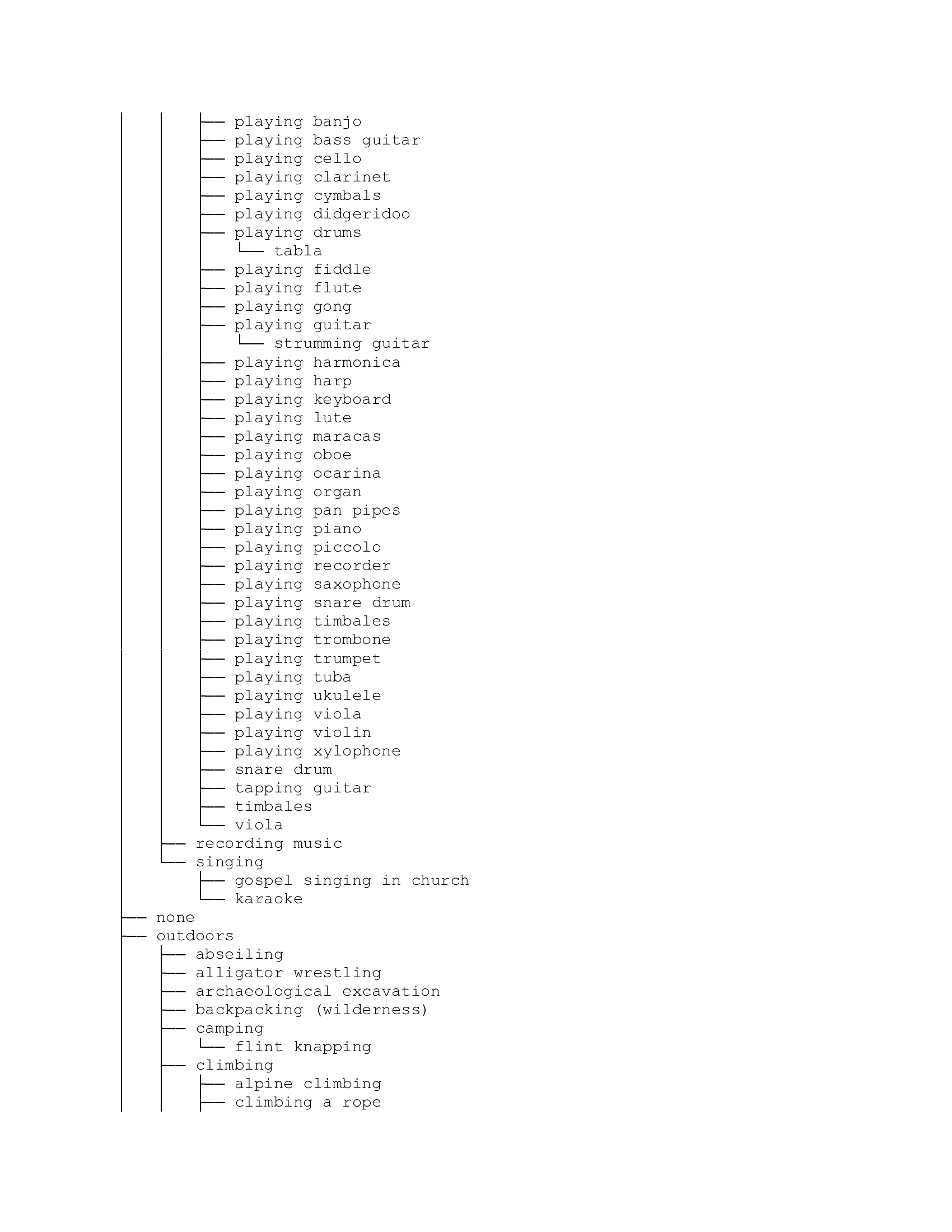}      \end{figure}
        \begin{figure}
        \centering
    \includegraphics[width=0.9\linewidth]{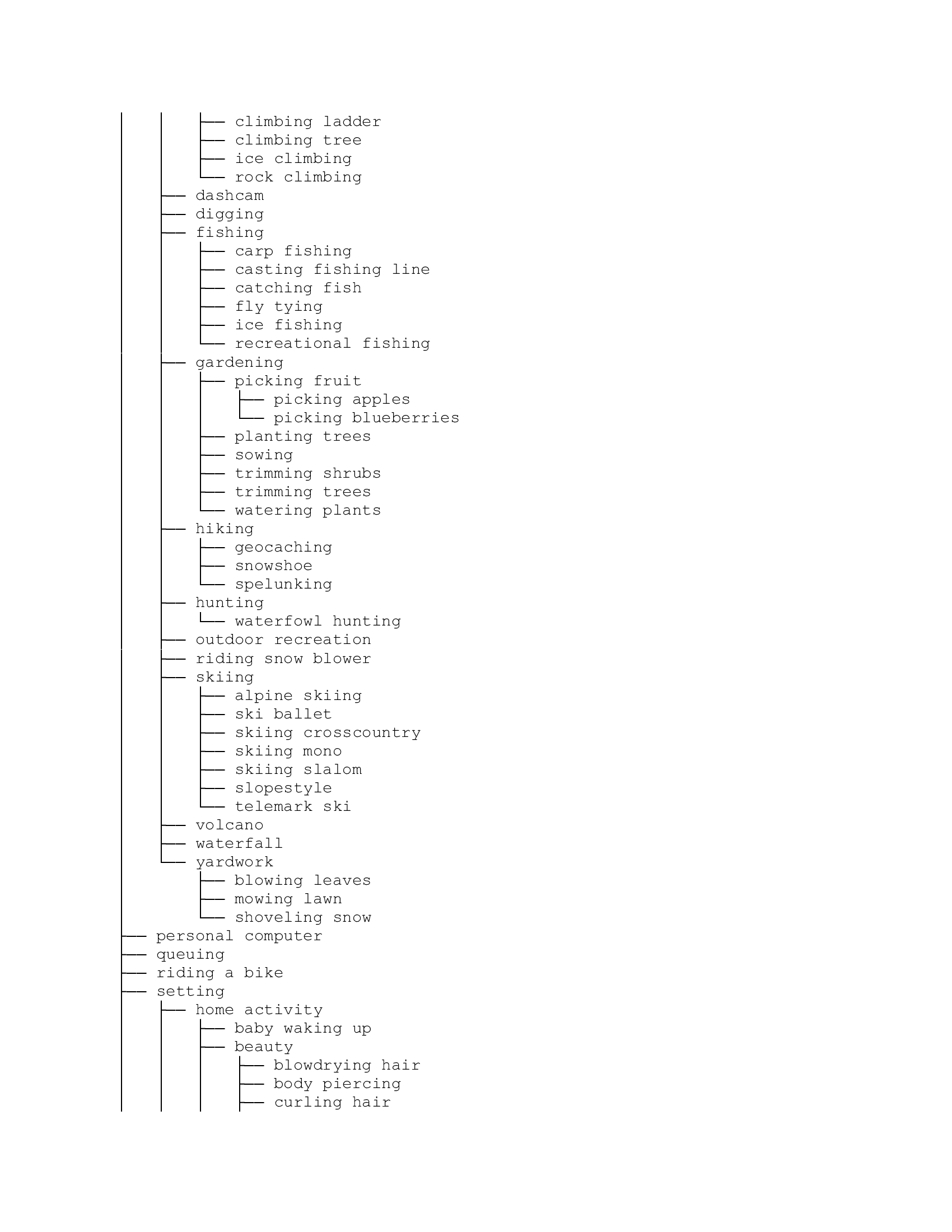}      \end{figure}
        \begin{figure}
        \centering
    \includegraphics[width=0.9\linewidth]{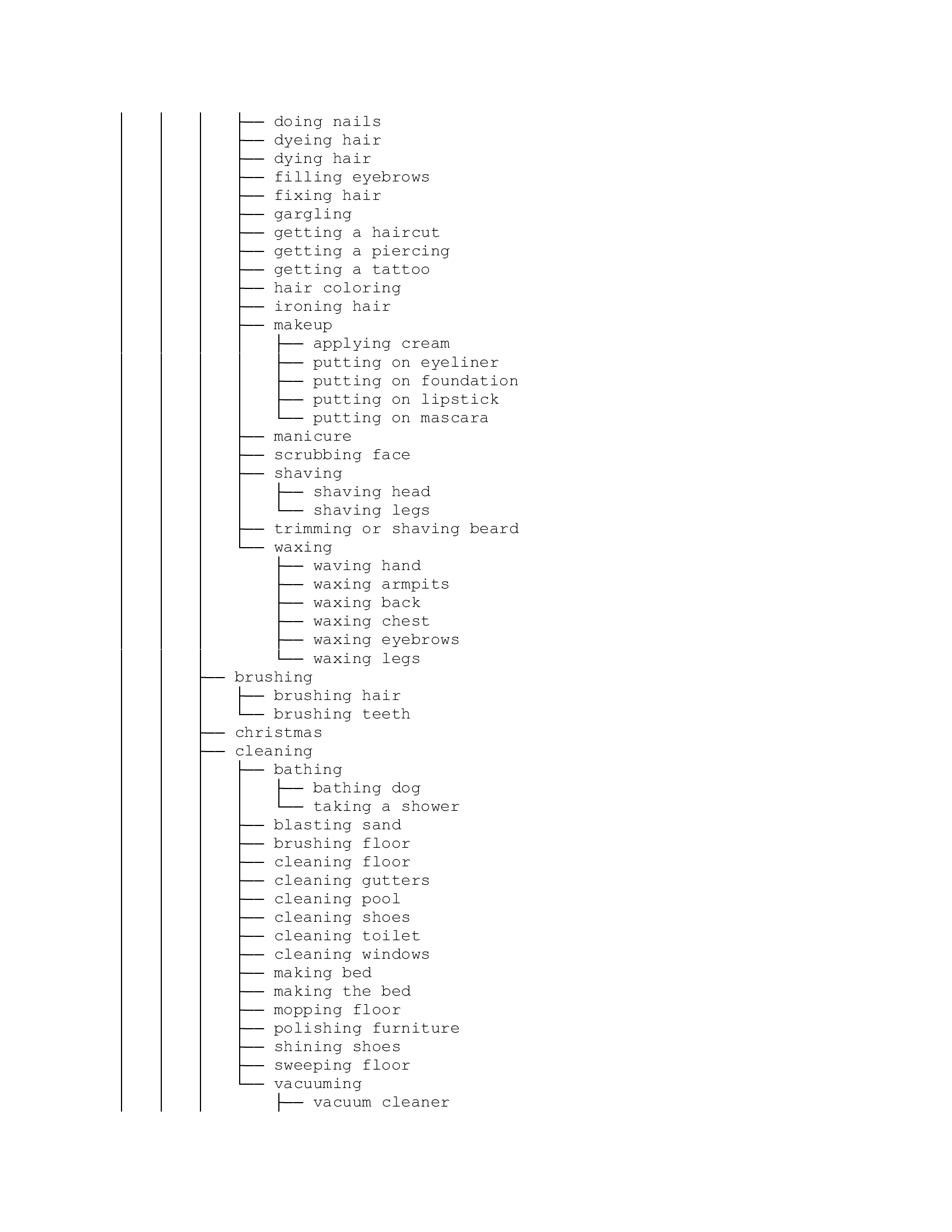}      \end{figure}
        \begin{figure}
        \centering
    \includegraphics[width=0.9\linewidth]{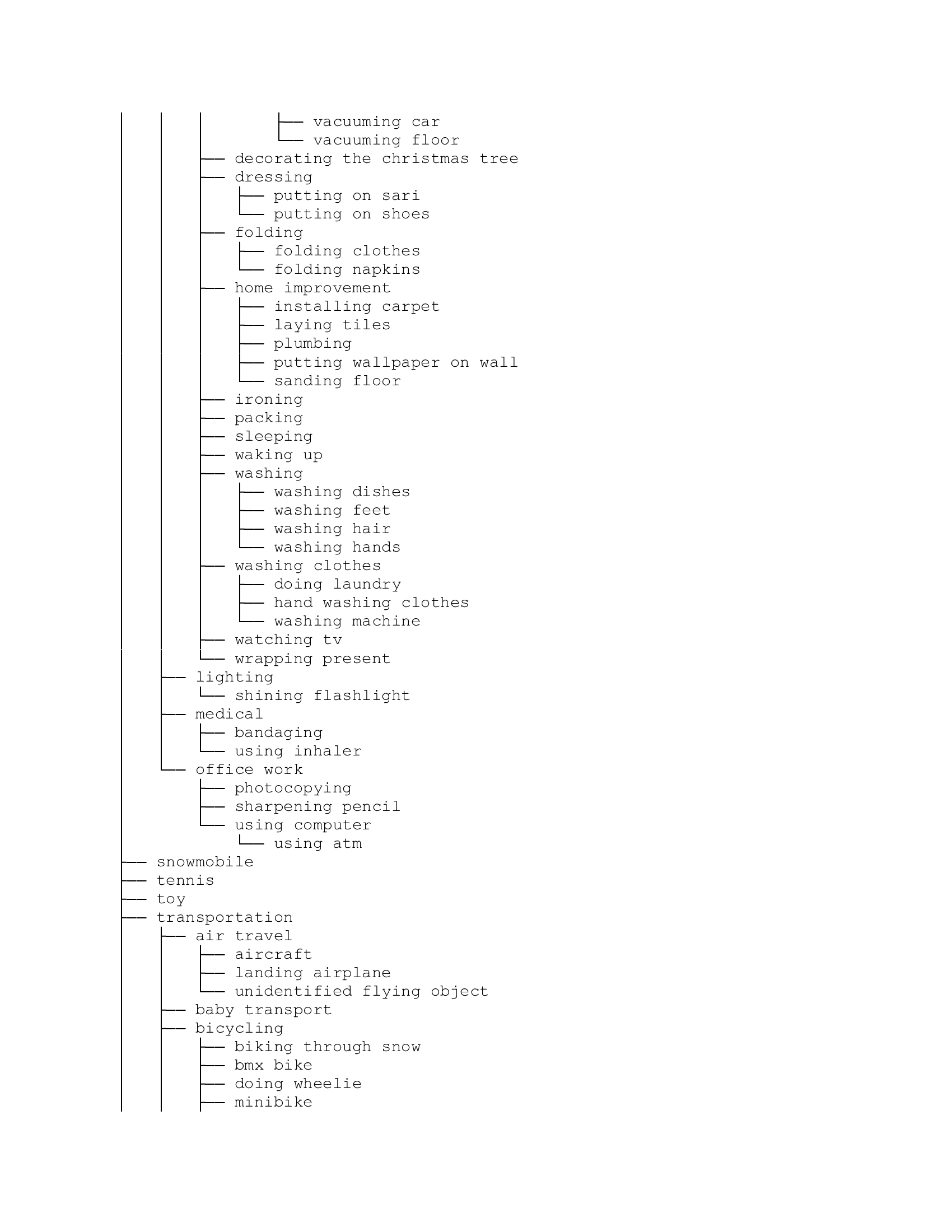}      \end{figure}
        \begin{figure}
        \centering
    \includegraphics[width=0.9\linewidth]{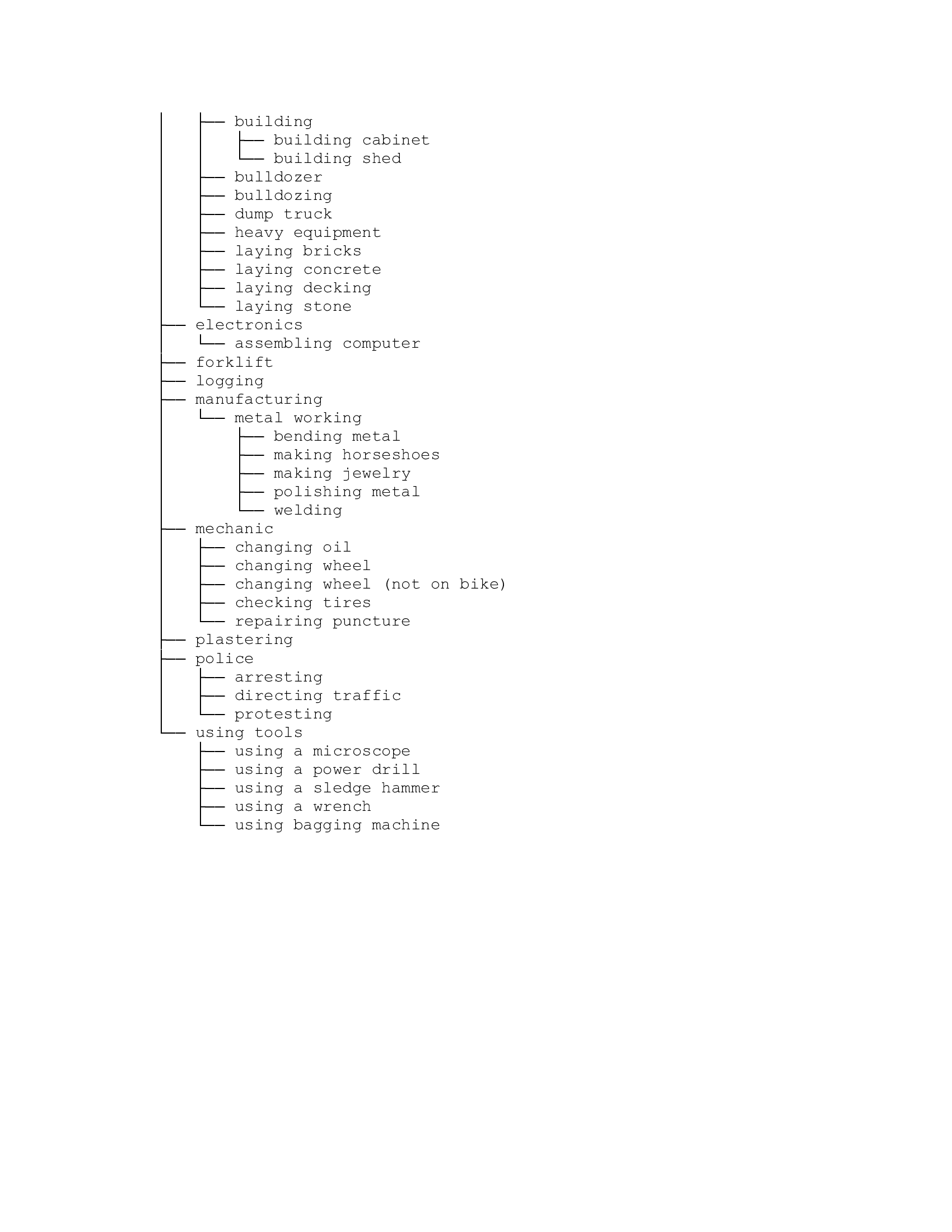}      \end{figure}
        \begin{figure}
        \centering
    \includegraphics[width=0.9\linewidth]{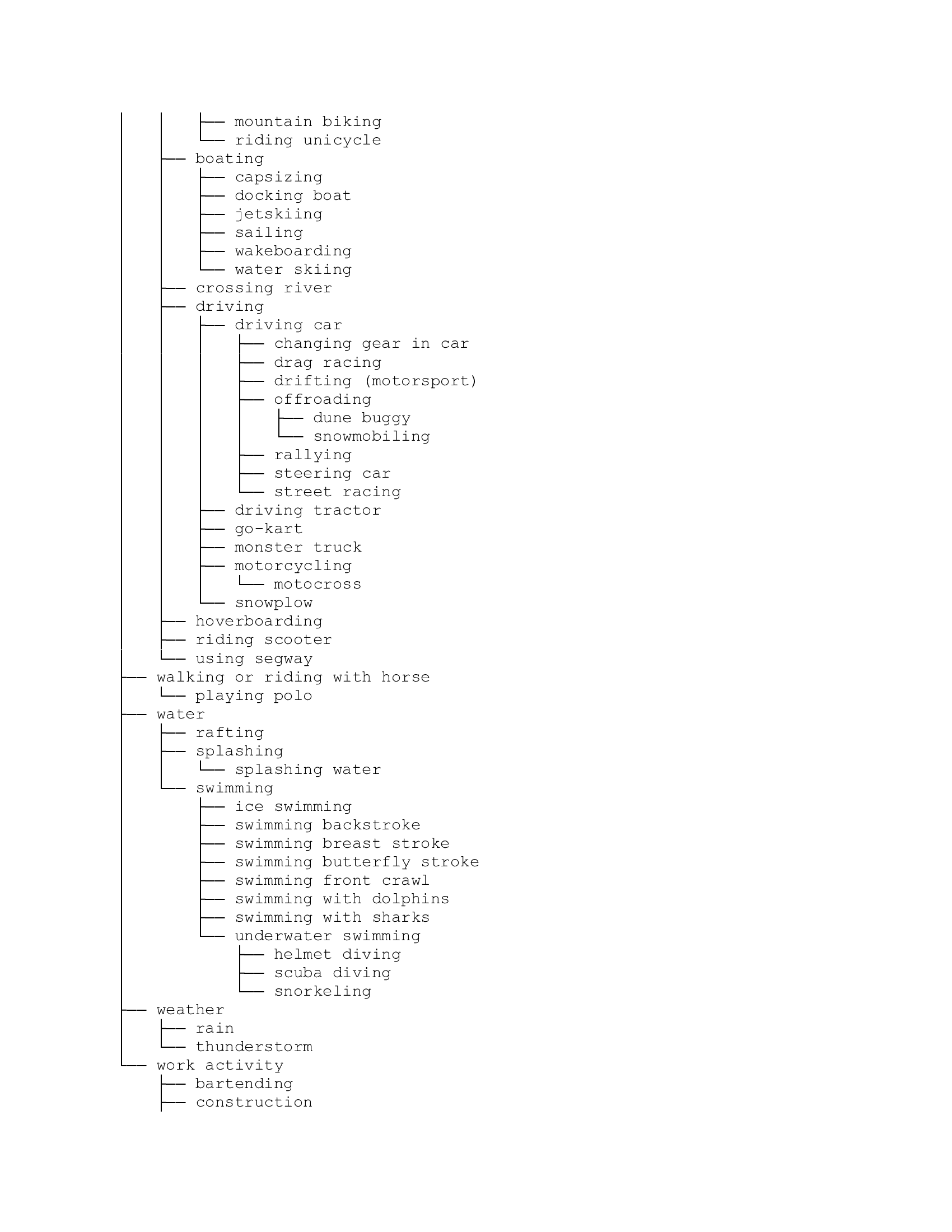}

    \caption{Full AViD hierarchy}
    \label{fig:hier}
\end{figure}

\end{document}